\documentclass[sigconf]{acmart}
\AtBeginDocument{%
  }

\setcopyright{acmlicensed}
\copyrightyear{2018}
\acmYear{2018}
\acmDOI{XXXXXXX.XXXXXXX}
\acmConference[Conference acronym 'XX]{Make sure to enter the correct
  conference title from your rights confirmation email}{June 03--05,
  2018}{Woodstock, NY}
\acmISBN{978-1-4503-XXXX-X/2018/06}
\usepackage{multirow}
\usepackage{subfigure,multirow,enumitem} 
\usepackage[ruled,vlined]{algorithm2e}
\usepackage{booktabs}
\usepackage{subcaption}
\usepackage{amsfonts}
\usepackage{colortbl}
\usepackage[table]{xcolor}
\definecolor{SereneMintBG}{HTML}{F7F9F9}
\definecolor{SereneMintPrimary}{HTML}{D6E8E4}
\definecolor{SereneMintAccent}{HTML}{B8D4E3}
\definecolor{SereneMintGray}{HTML}{A9A9A9}
\definecolor{SereneMintText}{HTML}{4A4A4A}
\begin{document}

\title{Effective Online 3D Bin Packing with Lookahead Parcels Using Monte Carlo Tree Search}

\author{Jiangyi Fang$^{1,4}$, Bowen Zhou$^{2}$, Haotian Wang$^{3}$, Xin Zhu$^{3}$, Leye Wang$^{1,4*}$}
\renewcommand{\authors}{Jiangyi Fang, Bowen Zhou, Haotian Wang, Xin Zhu Leye Wang}
\renewcommand{\shortauthors}{Jiangyi Fang, Bowen Zhou, Haotian Wang, Xin Zhu Leye Wang}
\affiliation{
\institution{$^{1}$ Key Lab of High Confidence Software Technologies (Peking University), Ministry of Education \country{China}}
\institution{$^{2}$ Faculty of Computing, Harbin Institute of Technology, Harbin, China}
\institution{$^{3}$ JD Logistic, Beijing, China}
\institution{$^{4}$ School of Computer Science, Peking University, Beijing, China}}\email{fangjiangyi2001@gmail.com, leyewang@pku.edu.cn}
\thanks{*Corresponding author}

\renewcommand{\shortauthors}{Trovato et al.}

\begin{abstract}
Online 3D Bin Packing (3D-BP) with robotic arms is crucial for reducing transportation and labor costs in modern logistics. While Deep Reinforcement Learning (DRL) has shown strong performance, it often fails to adapt to real-world short-term distribution shifts, which arise as different batches of goods arrive sequentially, causing performance drops. We argue that the short-term lookahead information available in modern logistics systems is key to mitigating this issue, especially during distribution shifts. We formulate online 3D-BP with lookahead parcels as a Model Predictive Control (MPC) problem and adapt the Monte Carlo Tree Search (MCTS) framework to solve it. Our framework employs a dynamic exploration prior that automatically balances a learned RL policy and a robust random policy based on the lookahead characteristics. Additionally, we design an auxiliary reward to penalize long-term spatial waste from individual placements. Extensive experiments on real-world datasets show that our method consistently outperforms state-of-the-art baselines, achieving over 10\% gains under distributional shifts, 4\% average improvement in online deployment, and up to more than 8\% in the best case--demonstrating the effectiveness of our framework.
\end{abstract}




\received{20 February 2007}
\received[revised]{12 March 2009}
\received[accepted]{5 June 2009}

\maketitle

\section{Introduction}
\label{sec:intro}
The global logistics market is anticipated to grow from 6.68 trillion to 11.27 trillion dollars in 2024 and 2033 with a growth rate of 5.98\% according to the market report~\cite{logistics_market_report}. For logistics companies (e.g., JD Logistics~\cite{jdlogistics} and Amazon~\cite{amazon}), bin packing represents a critical quality control measure within the supply chain. For example, improved space utilization of bin packing leads to substantial reductions in transportation costs~\cite{baldi2019generalized}. However, due to the high labor costs and practical sequential constraints (e.g., placement decisions must be made one by one at once for parcels traveling on a conveyor belt to prevent their accumulation), researchers are increasingly turning their research focus to online 3D bin packing using robotic arms~\cite{yang2023heuristics,wang2021dense}
\begin{figure}[t]
\centering
\includegraphics[width=0.43\textwidth]{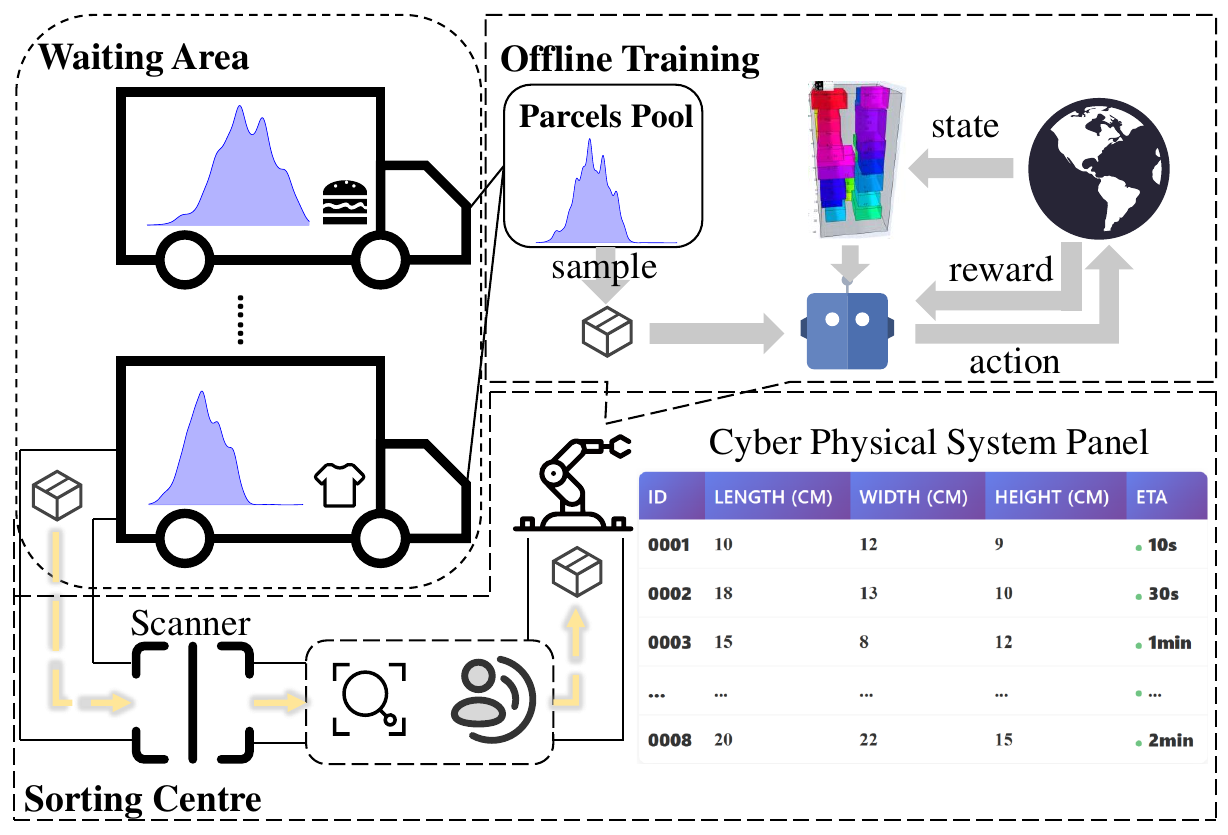} 
\vspace{-1.45em}
\caption{The mismatch between the real-world online scenario and offline DRL training pipeline.}
\label{fig: pitfall}
\vspace{-2.8em}
\end{figure}

Existing studies are primarily divided into heuristics~\cite{ha2017online,CrainicPT08,yang2023heuristics,karabulut2004hybrid,WangH19a,hu2017solving,zhao2021online,ojha2021generalized,Chazelle83,verma2020generalized} and Deep Reinforcement Learning (DRL) methods~\cite{verma2020generalized,zhao2021learning,zhao2021online,zhao2022learning,xiong2024gopt,yang2023heuristics}. Heuristics rules are usually summarized from workers' experience, e.g., layer first~\cite{Generalized}, deep bottom left first~\cite{KarabulutI04}, corner first~\cite{martello2000three} etc. Although DRL has recently surpassed heuristics after extensive training, both approaches fail to address a real-world scenario with short-term distributional shifts. As shown in Figure~\ref{fig: pitfall}, sorting centers with limited space must unload trucks sequentially in their arrival order. Since trucks carry parcels from different warehouses (e.g., food with varying sizes versus clothes with uniform sizes), this process creates a non-stationary parcel stream. However, current online DRL algorithms are trained on parcels sampled from a static pool~\cite{zhao2021online,zhao2022learning}, which aggregates parcels from all warehouses and is updated at a fixed cycle (e.g., weekly). Consequently, they cannot adapt to such shifts in practice.


With progress in Cyber-Physical Systems and sensing technology~\cite{oks2024cyber,zou2014radio,shao2021survey}, an online agent can acquire information about parcels as they pass through a scanner. As shown in Figure~\ref{fig: pitfall}, information shown on the panel forms a lookahead queue: newly scanned parcels are appended to the end, while parcels successfully packed into a bin are removed from the front. This queue partially reveals the size distribution of upcoming parcels, allowing the agent to detect shifts and adjust its policy in advance.


However, it is challenging to utilize lookahead information to enhance the online policy both efficiently and effectively. One approach is to use planning-based methods~\cite{huang2025milp,ali2022line,zhu2021learning} with sequential decision constraints. Nonetheless, their efficiency is often insufficient for online applications due to the inherent computational complexity (\textit{BFS} inference overhead over 100s shown in Table~\ref{tab: ablation_study}).  Another straightforward method, encoding lookahead parcels as part of the state in DRL context, proved insufficient by our experiments (\textit{PCT-lookahead} in Table~\ref{tab: main_results}) and existing works~\cite{zhao2021online}. Furthermore, \citeauthor{zhao2021online} proposed a reordering algorithm to condition the current parcel placement on subsequent ones~\cite{zhao2021online}. However, these approaches miss the fundamental issue. Given that RL is already forward-looking by nature~\cite{matsuo2022deep}, these methods fail to question why this inherent foresight is insufficient in online 3D-BP. In fact, lookahead parcels may not always bring about improvement to the same degree in the context of 3D-BP. As shown in Figure~\ref{fig: log_p_vs_Performance_gap} and Section~\ref{sec: theoretical_analysis}, the greater the deviation of the short-term distribution from the original, the larger the potential performance gain from lookahead information.

When faced with unfamiliar parcels, human workers may unpack recently placed items~\cite{song2023towards,yang2023heuristics} to conduct several placement trials, evaluating the best option based on their historical experience. Inspired by this human intelligence, we formulate the online 3D-BP with lookahead parcels as a Model Predictive Control (MPC) problem~\cite{kouvaritakis2016model,hansen2022temporal}. In this formulation, ``placement trial'' corresponds to a set of finite-horizon placement trajectories, while the ``evaluation through historical experience'' is achieved by using a well-trained online RL critic to select the optimal path. This mathematical formulation enables us to apply advanced techniques to tackle the problem. The main contributions of this paper are as follows:
\begin{itemize}[leftmargin=0.15cm]
    \item To the best of our knowledge, we are the first to establish that lookahead information is most valuable during short-term distributional shifts and to formulate the online 3D-PP with lookahead parcels as an MPC problem. Furthermore, we believe our approach could offer inspiration for other similar Markov Decision Process problems involving a finite lookahead horizon.
    
    \item We adapt the Monte Carlo Tree Search (MCTS)~\cite{swiechowski2023monte} framework to the 3D-BP scenario to solve the MPC problem formulated above. To adapt the MCTS exploration strategy for short-term shifts, we design a dynamic mechanism to adjust the exploration prior, balancing between a robust random policy and an offline-learned RL policy according to the likelihood of lookahead parcels appearing in the training dataset. Furthermore, to adapt the MCTS simulation and backup phases for the 3D-BP application, we introduce an auxiliary reward term that penalizes the wasted volume from each placement, considering its long-term effect on future parcels.
    
    \item We conduct extensive experiments on two real-world settings and an online evaluation, achieving over 10\% improvements against state-of-the-art DRL baselines under distributional shifts, and about 4\% average improvement in online deployment, validating the effectiveness of our proposed framework.
\end{itemize}


\section{Problem Formulation}
We will present our problem formulation in MDP form.

\noindent\textbf{State Space}. A state $s_t \in S$ is a tuple $s_t=(B_t,d_t)$, where $B_t$ can be represented by 2D height maps~\cite{zhao2021online} or packing configuration tree~\cite{zhao2022learning} and $d_t=(l_t,w_t,h_t)$ represents the length $l_t$, width $w_t$ and height $h_t$ of the parcel for step $t$.

\noindent\textbf{Action Space}. For a state $s_t$, the action space $A_{s_t}$ is the set of all valid placements for the parcel $d_t$. A placement $a_t = (x,y,z)\in A_{s_t}$ is a decision that maps to a specific position and orientation. Any valid action must satisfy all physical constraints (e.g., boundary, collision, stability).

\noindent\textbf{Transition Function}. The transition function $P(s_{t+1}\vert s_t,a_t)$ has a hybrid nature. The next state $s_{t+1}=(B_{t+1},d_{t+1})$ has two parts. For the next bin state $B_{t+1}$, the state transition function is \textbf{deterministic}, which is denoted as $B_{t+1} = f(B_t,d_t,a_t)$. For the next parcel state $d_{t+1}$, the state transition function is \textbf{stochastic}: $d_{t+1}\sim \mathcal{D} $. In practice, we collect an offline dataset $D$ to estimate the ideal distribution $\mathcal{D}$.

\noindent\textbf{Reward Function}. The immediate reward $r_t=R(s_t,a_t)$ can be defined as the volume of the packed item (i.e., $l_t*w_t*h_t$) normalized by the total bin volume (i.e., $L*W*H$ ), encouraging space utilization.

\noindent\textbf{Online 3D-BP with Lookahead Parcles}. Our problem setting includes several known components: a lookahead sequence of $N$ parcels (i.e., $d_t, d_{t+1}, \dots, d_{t+N-1}$ is known at step $t$), an offline-trained DRL critic $V_\theta$ and policy $\pi_\theta$, the offline dataset $D$, and a deterministic bin state transition function $f$. Given these components, our objective is to optimize the following function:
\begin{equation}
    \max_{a_t, \dots, a_{t+N-1}} \mathbb{E} \left[ \sum_{k=0}^{N-1} \left( R(s_{t+k},a_{t+k}) + V(s_{t+N})\right) \right]
\end{equation}
where $s_t = (B_t, d_t)$, $B_{t+1} = f(B_t, d_t, a_t)$, and $d_{t+N} \sim D$. We set the discount factor $\gamma=1$, as each episode terminates in a finite number of steps (i.e., the bin becomes full).
It is a typical MPC problem, allowing the agent to resist short-term distributional shifts by solving this optimization online. Following the receding horizon principle, once the optimal action trajectory is found, we execute only the first action $a_t^*$ and then replan at the subsequent step $t+1$ with new lookahead parcels appended.

\section{Theoretical Analysis}
\label{sec: theoretical_analysis}
In this section, we first present the optimization objective for the conventional online 3D-BP problem and introduce the basic MDP formulation as it applies to this problem. We then theoretically derive the relationship between the magnitude of a distributional shift and the corresponding performance degradation of a previously optimal policy.
\subsection{Preliminaries on online 3D-BP}
Given the definition above, the goal of conventional online 3D-BP  is to obtain an optimal policy $\bar{\pi}(a\vert s)$ which satisfies the following equation:
\begin{equation}
    \bar{\pi} = \arg \max\limits_\pi\sum\limits_{t=1}^\infty r_t
\end{equation}
whose corresponding optimal value function,$V_D^*(s)$, uniquely satisfies the Bellman optimality equation for this stationary environment:
\begin{equation}
    V_D^*(s_t) = \max\limits_{a\in A}(R(s_t, a_t) + \gamma \sum_{s_{t+1}\in S}P(s_{t+1}\vert s_t,a)V_D^*(s_{t+1}))
\label{eq:bellman_q_stationary}
\end{equation}
Since the transition function stochastic part only contains $d_{t+1}\sim D$, the equation can be rewritten as:
\begin{equation}
    V_D^*(s_t)=\max\limits_{a\in A}(R(s_t,a_t)+\gamma\mathbb{E}_{d_{t+1} \sim D} \left[ V_D^*(s_{t+1}) \right]
\end{equation}
\subsection{Failure in the non-stationary scenario.}

Considering a realistic non-stationary distribution, the overall long-term distribution $D$ is a mixture of these short-term distributions $D_x$:
\begin{equation}
D(d)=\sum\limits_xD_x(d)\cdot P(X=x)
\end{equation}
where $X$ denotes a discrete random variable of the type of each batch of parcels.

For a determinate batch $x$, the value function $V_{D_x}^{\bar{\pi}}$ in the new environment under policy $\bar{\pi}$ is defined as follow:
\begin{equation}
    V_{D_x}^{\bar{\pi}}(s_t)=\sum\limits_{a_t\in A} \bar{\pi} (a_t\vert s_t) \left(R(s_t,a_t)+\gamma\mathbb{E}_{d_{t+1} \sim D_x} \left[ V_{D_x}^{\bar{\pi}}(s_{t+1}) \right]\right)
\end{equation}
The optimal policy $\pi^*$ under the distribution $D_x$ is defined as follow:
\begin{equation}
    V_{D_x}^*(s_t)=\max\limits_{a_t\in A}\left(R(s_t,a_t)+\gamma\mathbb{E}_{d_{t+1} \sim D_x} \left[ V_{D_x}^*(s_{t+1}) \right]\right)
\end{equation}
The performance gap between the optimal policy $\pi^*$ and $\bar{\pi}$ under the distribution $D_x$ is defined as:
\begin{equation}
    \text{Gap}(s)=V_{D_x}^*(s)-V_{D_x}^{\bar{\pi}}(s)
\end{equation}
Actually, we can find that the performance gap can be bounded by the Total Variation distance between distributions.
\begin{equation}
    \lvert \text{Gap}(s)\rvert \leq \lvert V_{D_x}^*(s) - V_{D}^*(s)\rvert + \lvert V_{D_x}^{*}(s) - V_{D}^{\bar{\pi}}(s)\rvert \leq C \cdot D_{TV}(D_x,D)
\end{equation}
where $D_{TV}(D_x,D)$ is the Total Variation distance~\cite{bhattacharyya2022approximating,panaganti2022robust} between distributions and $C$ is a constant (proof details can refer to Appendix~\ref{sec: proof_details}). As a result, a larger distance between the short-term distribution $D_x$ and the long-term distribution $D$ can lead to a proportionally larger drop in performance, which also implies larger potential improvement (i.e., considering lookahead parcels).

\begin{figure}[htbp]
  \centering
  \includegraphics[width=0.85\linewidth]{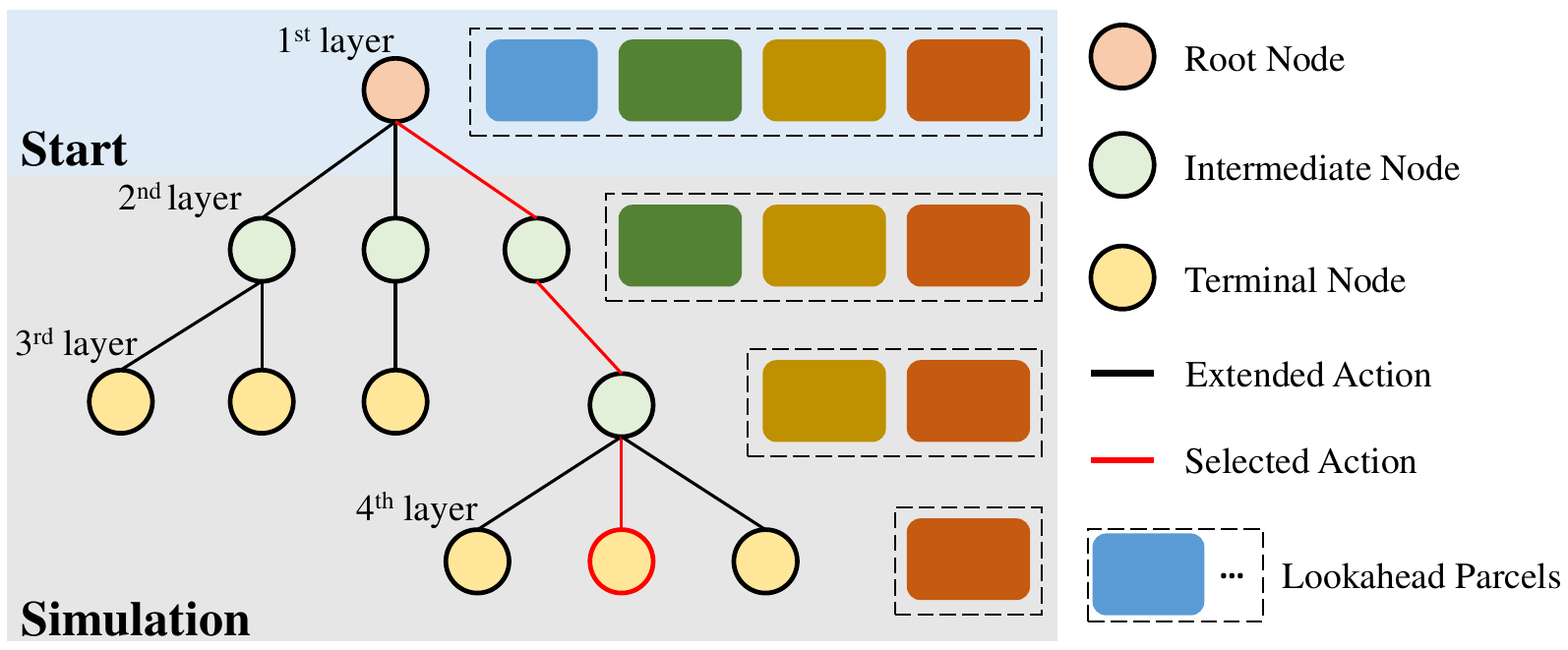}
  \vspace{-1em}
  \caption{Illustration for brute-force tree search solution for our MPC problem of online 3D-BP with lookahead parcels.}
  \label{fig: brute_force_search}
  \vspace{-1.75em}
\end{figure}

\section{Methodology}


\subsection{Tree Search Problem Modeling}

The MPC problem above can be viewed as a tree search problem. Figure~\ref{fig: brute_force_search} illustrates how to solve this problem at a single decision step $t$ using a brute-force search. The root node represents the initial state, comprising the current bin state $B_t$ and the first parcel $d_t$ from the lookahead queue. Starting from the root node (i.e., the $1^{st}$ layer), we simulate all possible placement actions $a \in A_{(B_t,d_t)}$ for the blue parcel. Each action, applied via the state transition function $f$, generates a new bin state $B'$ in the next layer. The edges in Figure~\ref{fig: brute_force_search} represent these state transitions. This process is repeated recursively; for instance, from each state in the second layer, we simulate placements for the green parcel to generate the third layer. The search terminates at a node (a terminal node) under two conditions: either the bin is full, or only one parcel remains in the lookahead sequence. In the latter case, we use the RL critic $V(s)$ to estimate the value of the terminal state. After the tree is fully explored, we identify the trajectory leading to the terminal node with the highest value (highlighted in red) and select the first action on this path as the optimal choice. Thus, solving the MPC problem is equivalent to finding the optimal root-to-leaf path in this search tree. Although effective, this brute-force search is computationally infeasible due to its exponential complexity of $O(\lvert A\rvert^N)$.
To address the inefficiency of brute-force search, we adapt the Monte Carlo Tree Search (MCTS) framework to the 3D-BP scenario to effectively balance exploration and exploitation. Our proposed \textit{MPC-3D-BP} framework retains the standard phases of MCTS: \textit{Selection}, \textit{Expansion}, \textit{Simulation}, and \textit{Backup}. Our main technical contributions lie in two phases: first, an adapted exploration strategy for the \textit{Selection} phase, and second, a novel penalty term for the value estimate used in the \textit{Simulation and Backup} phase. Both MCTS and our \textit{MPC-3D-BP} use the total number of trials $n$ to control the search time budget. 
\begin{figure}[htbp]
  \centering
  \includegraphics[width=.85\linewidth]{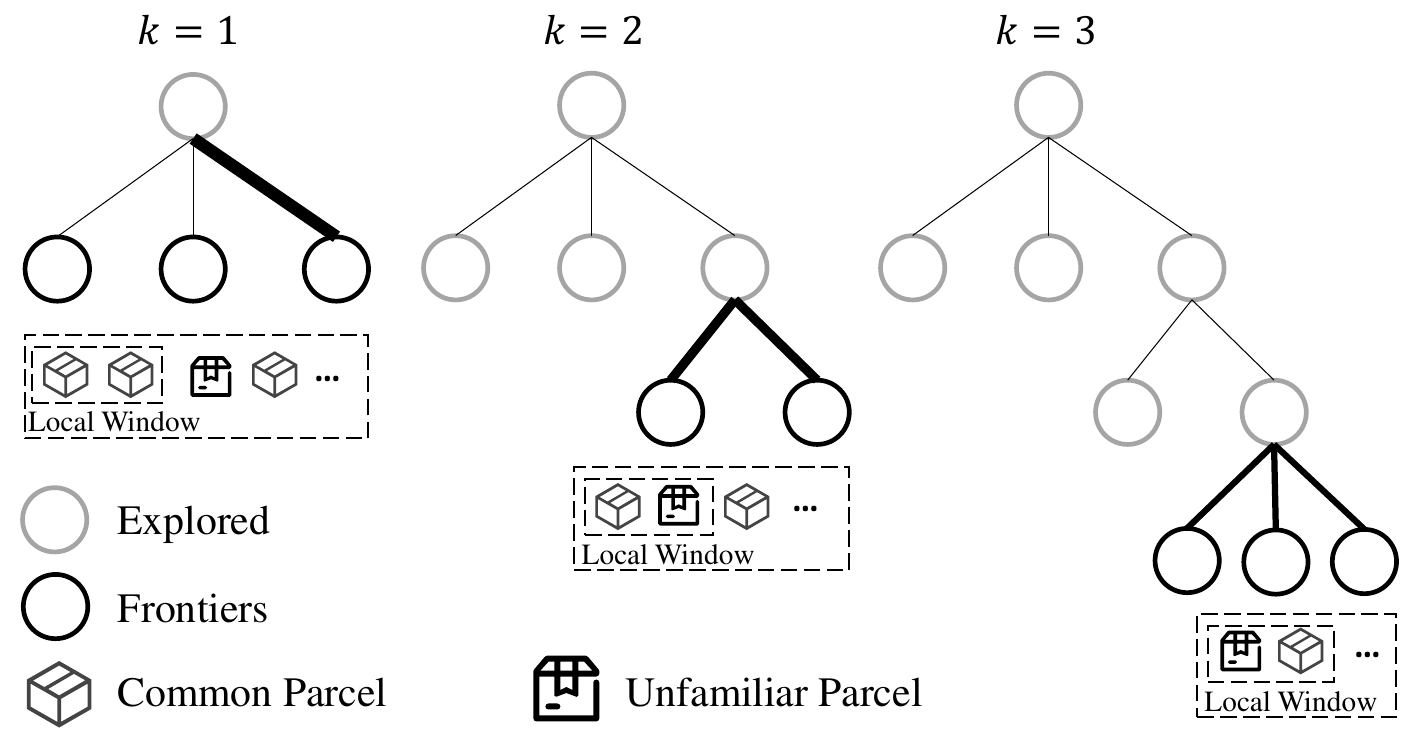}
  \vspace{-.75em}
  \caption{Step by step illustration for how Shift-Aware Polynomial Upper Confidence Trees guide the selection. The thickness of edges represents the prior of action probability.}
  \label{fig: SAUCB}
  \vspace{-2em}
\end{figure}
\subsection{MPC-3D-BP}
\subsubsection{Selection with Shift-Aware Polynomial Upper Confidence Trees}The goal of the \textit{Selection} phase in MCTS is to traverse the existing tree from the root to a promising leaf node for expansion. This process must strike a balance between the exploitation of nodes known to have high value and the exploration of less-visited nodes that may have high potential.

State-of-the-art MCTS variants, such as those used in AlphaZero~\cite{silver2018general}, employ the Polynomial Upper Confidence Trees (PUCT) algorithm for node selection. The action is chosen by maximizing:
\begin{equation}
    a^*=\arg\max\limits_a(Q(s,a)+c\cdot P_\pi(s,a)\cdot\frac{\sqrt{N(s)}}{1+N(s,a)})
\end{equation}
Here, $Q(s,a)$ is the exploitation term (mean action value), while the second term encourages exploration, guided by a prior probability $P_\pi(s,a)$ from a learned policy network and a fixed exploration coefficient $c$. $N(s)$ represents the total number of visits for node $s$ and $N(s,a)$ represents the number of visits for node $s$ 's child node generated through action $a$.

A key limitation of the standard PUCT formulation is its reliance on the learned policy prior $P_\pi(s,a)$ being equally reliable in all situations. This assumption is violated in our setting when a short-term distributional shift occurs. For a rare and unseen lookahead sequence $I_{sub}$, the learned prior $P_\pi(s,a)$ can be misleading, yet the PUCT algorithm continues to trust it. This can trap the search in suboptimal regions suggested by an unreliable``intuition".

To address this, we propose a novel mechanism that dynamically modulates the exploration prior based on the rarity of the current lookahead sequence. Instead of using the raw policy prior $P_\pi(s,a)$, we introduce a dynamic shift-aware prior, $P_{SA}(a\vert s)$, which performs a principled interpolation between trusting the learned policy and reverting to a robust, uniform exploration strategy.

The dynamic prior is defined as:
\begin{equation}
    P_{SA}(a\vert s) = \alpha(s)\cdot P_\pi(s,a) + (1-\alpha(s))\cdot \frac{1}{\lvert A_s\rvert}
\end{equation}
The interpolation weight $\alpha(s)\in[0,1]$ is the``Familiarity Score'' of the local lookahead subsequence $I_{k}$ at state $s$. The calculation of $\alpha(s)$ is designed to be simple yet effective. The process is divided into two stages.

\noindent\textbf{Offline Pre-computation.} We process the entire training dataset $D$ to build a probability look-up table for item types. First, we discretize the continuous item dimensions $(l,w,h)$ into a finite vocabulary of $M$ unique item types using heuristics through visualization. Then, we compute the empirical probability $P(type_j)$ for each item type by counting its frequency in the training data.

\noindent\textbf{Online Calculation.} During the MCTS search, for the given node $s$ at the $k^{th}$ layer of the tree, we calculate its Familarity Score $\alpha(s)$ as follows. We consider a fixed-size (empirically set to $3$ for the pattern 2 normal parcels and 1 big one in Figure~\ref{fig: case_study}) local window of $w$ items, starting from the current item to be placed in the node's lookahead queue, denoted as $I_k=(d_{t+k-1},d_{t+k},...,d_{t+k+w-1})$. Assuming the occurrence of each item in this window is independent, the joint probability is the product of their probabilities, which are retrieved from our pre-computed table. The Familiarity Score is this joint probability:
\begin{equation}
    \alpha(s) = P(I_k)=\prod_{i=1}^wP(Disc(d_{t+i+k-1})
\end{equation}
where $Disc(d_{t+i+k-1})$ is the function that maps item $d_{t+i+k-1}$ to its discrete type.

By substituting this dynamic prior into the PUCT formula, we get our final \textbf{Shift-Aware PUCT} selection rule:
\begin{equation}
    a^*=\arg\max\limits_a(Q(s,a)+c\cdot P_{SA}(a\vert s)\cdot\frac{\sqrt{N(s)}}{1+N(s,a)})
\end{equation}
This formula exhibits intelligent, adaptive behavior, being aware of the rarity of the local lookahead sequence.
\begin{itemize}[leftmargin=0.2cm]
    \item \textbf{When the local lookahead sequence is common}: The prior term relatively approximates $P_\pi(a\vert s)$. As shown in the $k=1$ step in Figure~\ref{fig: SAUCB}, the algorithm trusts its learned intuition and efficiently exploits its knowledge, behaving like a standard PUCT (i.e., the distribution of actions with little uncertainty).
    \item \textbf{When a significant distributional shift occurs}: The prior term approximates the uniform distribution $\frac{1}{\vert A_s\rvert}$. As shown in the $k=2,k=3$ step in Figure~\ref{fig: SAUCB}, the algorithm recognizes that its learned intuition is unreliable and gracefully degrades to a robust, non-informative exploration strategy, preventing the search from being led astray by a misleading prior.
\end{itemize}
This mechanism allows the search to dynamically allocate its exploration preference, investing more in broad exploration for novel challenges while efficiently leveraging learned knowledge for familiar situations.

\subsubsection{Expansion.}
Once the \textit{Selection} phase reaches a leaf node $s_L$ that has not been expanded before, the Expansion phase is initiated to grow the search tree. To avoid the intractable cost of creating nodes for all possible actions, we follow state-of-the-art methods and use our learned policy network $P_\pi(a\vert s)$ to guide this process.

For each valid action $a\in A_{s_L}$, we add a corresponding new child node to the tree. The edge leading from the parent node $s_L$ to this new child is initialized with the following statistics:

\begin{itemize}
    \item Visit Count: $N(s_L,a)\leftarrow0$
    \item Action Value: $Q(s_L,a)\leftarrow0$
    \item Prior Probability: $P_{SA}(a\vert s)\leftarrow\alpha(s)\cdot P_\pi(s,a) + (1-\alpha(s))\cdot \frac{1}{\lvert A_s\rvert}$
\end{itemize}

After this step, the node $s_L$ is now an internal node, and the search tree has been successfully expanded with new leaves ready for future selection and evaluation.
\subsubsection{Simulation and Value Backup}
\label{sec:method_backup}
\begin{figure}[htbp]
  \centering
  \includegraphics[width=.85\linewidth]{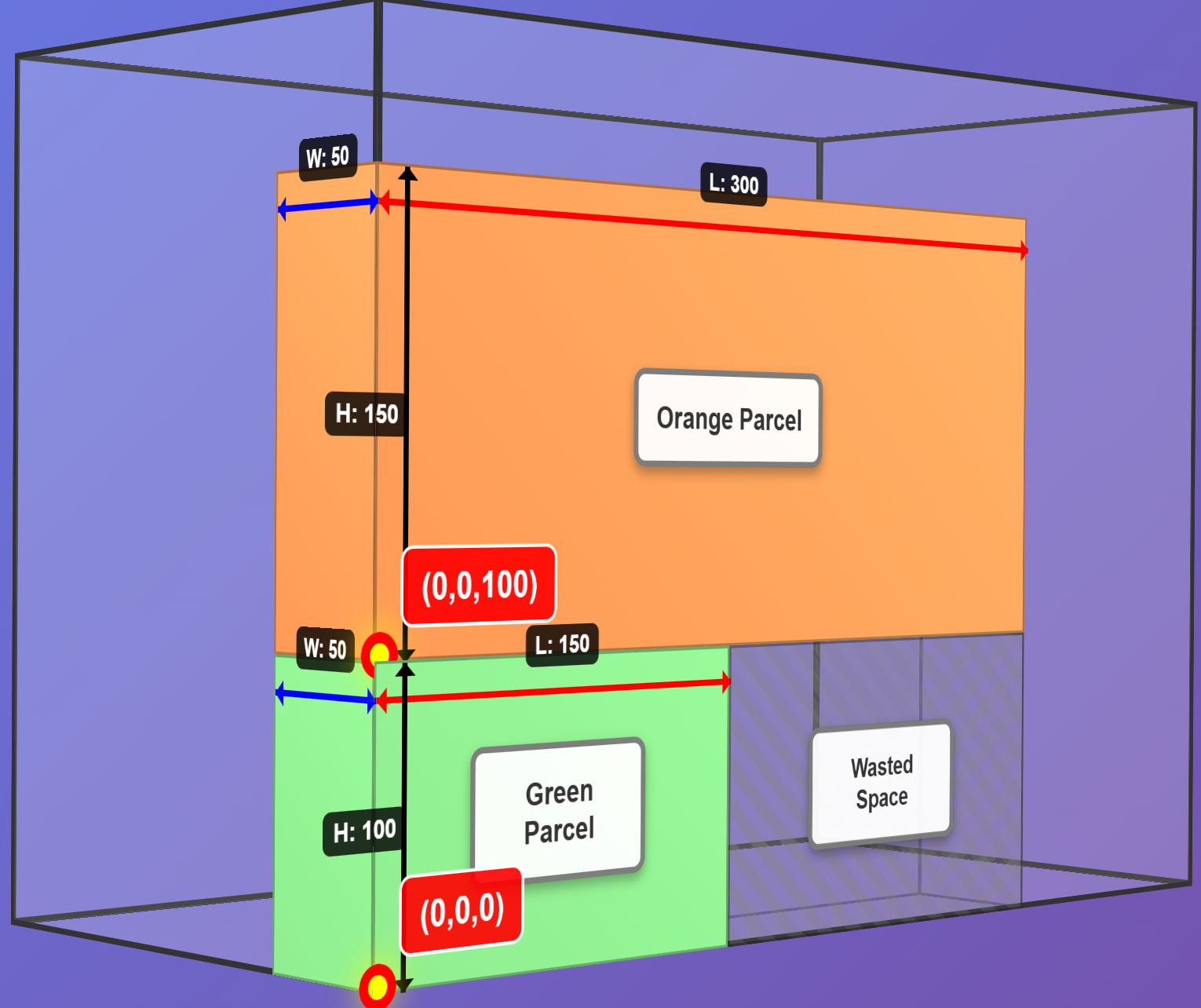}
  \vspace{-.75em}
  \caption{Illustration for waste space calculation.}
  \label{fig: waste_space}
  \vspace{-1em}
\end{figure}
The \textit{Simulation} phase in traditional MCTS involves performing a random or heuristic-based rollout from a newly expanded leaf node to the end of the game to obtain an estimate of its value. However, to improve efficiency, modern implementations like AlphaZero~\cite{silver2018general} often replace this costly rollout with a direct value estimate from a learned critic, $V_\theta$. In our MPC-based framework, the value of an entire $N$-step plan (a root-to-leaf path in the search tree) is similarly backed up. The standard approach is to sum the immediate rewards collected along the path and bootstrap the terminal value using the critic:
\begin{equation}
    Q(a_t,a_{t+1},...a_{t+N-1}) = \sum_{k=0}^{N-1} r_{t+k} + V_\theta(s_{t+N})
\label{eq:base_backup}
\end{equation}
The rationale for explicitly summing the short-term rewards, rather than just relying on $V_\theta(s_t)$, is to ground the value estimate in the factual outcomes of the known lookahead sequence, mitigating the potential unreliability of the critic when facing a short-term distributional shift.

However, this standard backup mechanism has a significant drawback in the context of 3D-BP. The standard reward signal, $r_{t+k}$ (typically based on placed volume), is often insensitive in the early and middle stages of the packing process. Most valid placements yield similar, positive rewards, failing to differentiate a structurally efficient placement from one that catastrophically wastes space for future items. This places an enormous burden on the terminal value estimate $V_\theta(s_{t+N})$, which may not be robust enough to distinguish between states that have subtle but critical structural differences.

Therefore, inspired by the physical constraints of robotic packing, we propose a heuristic reward shaping scheme to create a more discerning, step-wise reward signal. Instead of only rewarding placed volume, we directly penalize the creation of wasted space along the short-term trajectory. In realistic scenarios, robotic arms can only place items from the top down due to gravity. Consequently, any space left underneath a newly placed item becomes practically inaccessible for all future placements and is effectively wasted. As shown in Figure~\ref{fig: waste_space}, placing the orange parcel on top of the green one creates wasted space that is inaccessible to any future parcels.

To measure this, we introduce the \textbf{Wasted Space Penalty}. This penalty, $R_p(s_k,a_k)$, is calculated as the normalized volume of the space directly beneath the footprint of the item placed by action $a_k$. For example, in Figure~\ref{fig: waste_space}, the wasted ratio is
\begin{equation}
    \begin{aligned}
        R_p(s_k,a_k)&=(l_o*w_o-l_g*w_g)*(z_o-z_g)/(L*W*H)\\
    &=(150*50-300*50)*(100-0)/(400*200*300)\\
    &=3.125\%
    \end{aligned}
\end{equation}
where $l_o, w_o, h_o$ are the dimensions of the orange parcel, $l_g, w_g, h_g$ are the dimensions of the green one, and $z_o, z_g$ are the placement coordinates of the orange and green parcels, respectively. This volume represents space that can no longer be utilized due to gravitational placement constraints. Our new shaped reward function, $r'$, incorporates this penalty:
\begin{equation}
    r'_{t+k} = R_{\text{vol}}(s_k,a_k) - \lambda \cdot R_p(s_k,a_k)
    \label{eq:shaped_reward}
\end{equation}
where $R_{\text{vol}}$ is the standard volume-based reward (e.g., the $R_{\text{vol}}$ for orange parcel's placement in Figure~\ref{fig: waste_space} is calculated as $\frac{l_o*w_o*h_o}{L*W*H}=\frac{300*50*150}{400*200*300}=9.375\%$), and $\lambda$ is a hyperparameter to adjust the weight of penalty term (we set $\lambda=0.5$ due to Figure~\ref{fig: lambda}).

The final value that is backed up through the MCTS tree, which we denote as $Q'$, uses this enriched reward signal:
\begin{equation}
    Q'(a_t,a_{t+1},...a_{t+N-1}) =  \sum_{k=0}^{N-1}  r'_{t+k}  +  V_\theta(s_{t+N})
    \label{eq:bhec_backup}
\end{equation}
After calculating the value $Q'$, the algorithm backs up this value along the selected trajectory, updating the $Q(s,a)$ value for each node visited.

\section{Offline Evaluation}

\subsection{Experiment Setup}
\subsubsection{Datasets, baselines, and experiment settings} 
We obtain the ground-truth parcel stream by mining historical logs and set the lookahead number to $N=4$, as this covers most online cases (see Figure~\ref{fig: lookahead} in Appendix~\ref{sec: illustrative_example_data_evidence}). To evaluate the effectiveness of our framework, we compare it against four heuristic methods (\textit{LSAH}~\cite{hu2017solving}, \textit{OnlineBPH}~\cite{ha2017online}, \textit{MACS}~\cite{HuXCG0020}
, \textit{DBL}~\cite{KarabulutI04}) and two DRL methods (\textit{PCT}~\cite{zhao2022learning}, \textit{GOPT}~\cite{xiong2024gopt}) as baselines. Additionally, we implement two variants that incorporate lookahead information, \textit{PCT-lookahead} and \textit{PCT-reorder}~\cite{zhao2021online}. To validate our specific technical contributions, we also conduct an ablation study with several alternative techniques (\textit{BFS}~\cite{zhou2006breadth}, \textit{Random}~\cite{pemantle1995critical}, \textit{RTDP}~\cite{barto1995learning}, \textit{MCTS}~\cite{silver2018general}). Additionally, we test the effectiveness of our framework under two settings \textbf{Realistic} and \textbf{Virtual}. We set up the \textbf{Virtual} setting following the setting in \textit{PCT}~\cite{zhao2022learning}, while the \textbf{Realistic} setting contains more real-world constraints on action space like collision avoidance.  Details of datasets, baselines, and settings are provided in Appendix~\ref{sec: implementation_details}. 
\subsubsection{Evaluation metrics}
According to~\citeauthor{zhao2021online}, we evaluate the performance of the 3D-BP algorithms using two core metrics~\cite{zhao2021online}: space utilization (``Space Uti.'') and the number of packed items (``\# Items'') (in Appendix~\ref{sec: metric}). To assess efficiency, we also record the average inference time per placement, denoted as ``Infer. Time''. Furthermore, we define two evaluation scenarios: \textbf{Overall} and \textbf{Shift}. The \textbf{Overall} scenario computes these metrics across all test bins, whereas the \textbf{Shift} scenario calculates them exclusively for bins where the incoming parcel stream exhibits a significant distributional shift.
 
\subsection{Results and Analysis}
\subsubsection{Main Results.}
\begin{table*}[t]
\small
\caption{Results on the same dataset for Realistic and Virtual settings. DRL baselines are colored with light green. The best two results are highlighted (\textbf{\textit{best}} is in bold and \textit{italic}, \textbf{second-best} is in bold).}
\setlength{\tabcolsep}{1.4mm}{
\begin{tabular}{lcccccccccccccccc}
\toprule
& \multicolumn{5}{c}{\textbf{Realistic}} & \multicolumn{5}{c}{\textbf{Virtual}} \\ 
 \cmidrule(lr){2-6} \cmidrule(lr){7-11}
& \multicolumn{2}{c}{\textbf{Overall}} & \multicolumn{2}{c}{\textbf{Shift}} & \multirow{2}{*}{Infer. Time} & \multicolumn{2}{c}{\textbf{Overall}} & \multicolumn{2}{c}{\textbf{Shift}} & \multirow{2}{*}{Infer. Time} \\
\cmidrule(lr){2-3} \cmidrule(lr){4-5} \cmidrule(lr){7-8} \cmidrule(lr){9-10}
& Space Uti. & \# Items & Space Uti. & \# Items &  & Space Uti. & \# Items & Space Uti. & \# Items & \\
\midrule
\multicolumn{3}{l}{\textbf{Baselines}} \\
\textit{LSAH}  &  0.2396  & 15.29 & 0.2570 & 21.20 & $<1s$ & 0.2928  & 19.47 & 0.3218 & 23.64 & $<1s$ \\
\textit{OnlineBPH}& 0.2903  & 18.50 & 0.3345 &  27.07 & $<1s$ & 0.3568  & 23.28 & 0.3700 & 26.83 & $<1s$ \\
\textit{MACS} & 0.3826  & 24.41 & 0.3899 & 31.44 & $<1s$ & 0.4783  & 30.68 & 0.4668 & 32.86 & $<1s$ \\
\textit{DBL} &  0.3677  & 23.56 & 0.4234 & 33.90 & $\sim5s$ & 0.4379  & 28.37 & 0.4584 & 32.46 & $\sim5s$ \\
\rowcolor{SereneMintPrimary}\textit{PCT} & 0.5395  & 34.14 & 0.4600 & 36.37 & $<1s$ & 0.5519  & 35.30 & 0.5227 & 36.47 & $<1s$ \\
\rowcolor{SereneMintPrimary}\textit{GOPT} & 0.5211  & 33.14 & 0.4461 & 36.06 & $<1s$ & 0.5331  & 34.27 & 0.5234 & 36.60 & $<1s$ \\
\midrule
\multicolumn{3}{l}{\textbf{Lookahead Information Enhanced}} \\
\textit{PCT-lookahead} & 0.5286  & 33.41 & 0.4788 & 37.59 & $<1s$ & 0.5457  & 34.90 & 0.5428 & 37.63 & $<1s$ \\
\textit{PCT-reorder} & \textbf{0.5435}  & \textbf{34.39} & 0.4903 & 38.36 & $\sim3s$ & \textbf{0.5638}  & \textbf{36.01} & 0.5651 & 38.98 & $\sim3s$ \\
\midrule
\multicolumn{3}{l}{\textbf{Ours}} \\
\textit{MPC-PCT}  & \textbf{\textit{0.5498}}  & \textbf{\textit{34.80}} & \textbf{\textit{0.5258}} & \textbf{40.61} & $\sim10s$ & \textbf{\textit{0.5765}}  & \textbf{\textit{36.83}} & \textbf{\textit{0.6034}} & \textbf{\textit{41.45}} & $\sim10s$ \\
\textit{MPC-GOPT}  & 0.5403  & 34.31 & \textbf{0.5165} & \textbf{40.10} & $\sim20s$ & 0.5527  & 35.48 & \textbf{0.5927} & \textbf{40.70} & $\sim20s$ \\
\bottomrule
\end{tabular}}
\label{tab: main_results}
\vspace{-1.2em}
\end{table*}

In Table~\ref{tab: main_results}, we report the Space Uti., \# Items and Infer. Time in both \textbf{Overall} and \textbf{Shift} scenarios under two settings. Table~\ref{tab: main_results} shows that methods properly considering lookahead parcels (\textit{PCT-reorder} and \textit{MPC-PCT}) achieve better results than baselines, justifying the effectiveness of lookahead information. Specifically, \textit{MPC-PCT} achieves over 15\% improvement against \textit{PCT} in the \textbf{Shift} scenario on both metrics, indicating that lookahead parcels are especially useful under distributional shifts. Besides, our \textit{MPC-PCT} achieves notably better results than \textit{PCT-reorder} and \textit{PCT-lookahead} (over 10\% improvement in the \textbf{Shift} scenario on both metrics), indicating our method of considering lookahead parcels reaching the state-of-the-art in practice. Besides, we also test our framework on another DRL baseline \textit{GOPT} and obtain similar observations (\textit{MPC-GOPT} and \textit{MPC-PCT} obviously outperform the corresponding baselines with the green background color), which validates our framework's generalization ability. Additionally, we can observe all results above under the \textbf{Virtual} setting, indicating our framework generalization across different action spaces (with different practical constraints). Although our framework achieves promising results, it takes about 3 times higher inference costs than \textit{PCT-reorder} and over 10 times higher inference costs than \textit{PCT}. However, it can meet the online requirements according to the detailed analysis in Section~\ref{sec: online_deployment}.

\subsubsection{Ablation study.}
\begin{table}[t]
\small
\caption{Ablation study results under the Realistic setting. The best two results excluding \textit{BFS} are highlighted (\textbf{\textit{best}} is in bold and \textit{italic}, \textbf{second-best} is in bold).}
\setlength{\tabcolsep}{1.3mm}{
\begin{tabular}{lccccc}
\toprule
& \multicolumn{5}{c}{\textbf{Realistic}} \\ 
\cmidrule(lr){2-6} 
&  \multicolumn{2}{c}{\textbf{Overall}} & \multicolumn{2}{c}{\textbf{Shift}} & \multirow{2}{*}{Infer. Time}\\
\cmidrule(lr){2-3} \cmidrule(lr){4-5}
&Space Uti. & \# Items &Space Uti. & \# Items & \\
\midrule
\multicolumn{3}{l}{\textbf{Alternative Techniques}}  \\
\textit{BFS}  & 0.5513 & 34.91 & 0.5403 & 41.71 & $\sim100s$ \\
\textit{Random} & 0.5436  & 34.42 & 0.4955
 & 38.75 &  $\sim15s$\\
\textit{RTDP} & 0.5453
  & 34.50
 & 0.4956
 & 38.54 & $\sim18s$\\
\textit{MCTS} & 0.5402
  & 34.19
 & 0.4679
 & 37.03 & $\sim10s$\\
\midrule
\multicolumn{2}{l}{\textbf{Ours}} \\
\textit{$\sim$ w/o WSP}  & \textbf{0.5494}  & \textbf{34.77} & \textbf{0.5218} & \textbf{40.32} & $\sim10s$ \\
\textit{$\sim$ w/o SAPUCT}  & 0.5410  & 34.25 & 0.4723 & 37.41
 & $\sim10s$\\
\textit{MPC-PCT}  & \textbf{\textit{0.5498}}  & \textbf{\textit{34.80}} & \textbf{\textit{0.5258}} & \textbf{\textit{40.61}} & $\sim10s$ \\
\bottomrule
\end{tabular}}
\label{tab: ablation_study}
\vspace{-1.7em}
\end{table}
To better evaluate our technical contributions, we conduct an ablation study by implementing several alternative tree search techniques to solve our MPC problem. Table~\ref{tab: ablation_study} displays the results under the Realistic setting. First, all tree search techniques outperform the baseline \textit{PCT} in both scenarios, which validates the effectiveness of framing the problem within an MPC framework. Second, while brute-force search (\textit{BFS}) achieves the best performance, its inference overhead is approximately 10 times higher than that of the other techniques, making it impractical. Additionally, \textit{RTDP} with its $\epsilon$-greedy strategy performs more robustly than the pure exploration of the \textit{Random} baseline in the \textbf{Overall} scenario. However, a standard \textit{MCTS} with Polynomial Upper Confidence Trees relies too heavily on the prior DRL policy, resulting in limited performance gains. Notably, removing our proposed exploration strategy (\textit{$\sim$ w/o SAPUCT}) causes a significant performance drop, particularly in the \textbf{Shift} scenario (an 11.3\% reduction in space utilization). This result underscores the importance of a well-designed strategy that balances the DRL prior with random exploration. Similarly, the underperformance of the variant without our waste space penalty (\textit{$\sim$ w/o WSP}) confirms the effectiveness of this proposed term.

\subsubsection{Robustness analysis at different levels.}
\newcommand{\fourfigcol}{-2mm}
\begin{figure}[t]
\centering
\subfigure[Top Level]{
\includegraphics[width=0.23\textwidth]{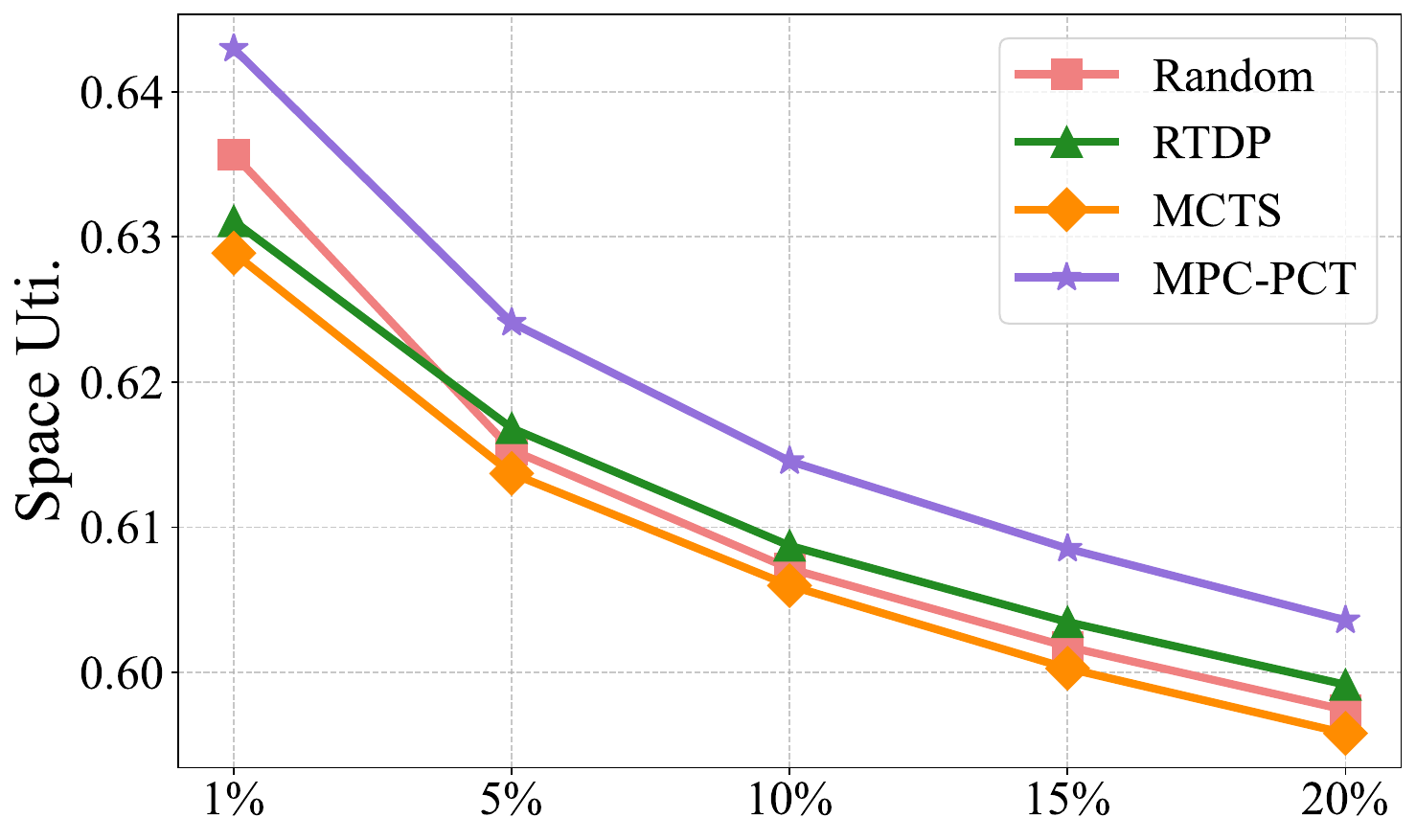}
}\hspace{\fourfigcol}
\subfigure[Bottom Level]{
\includegraphics[width=0.23\textwidth]{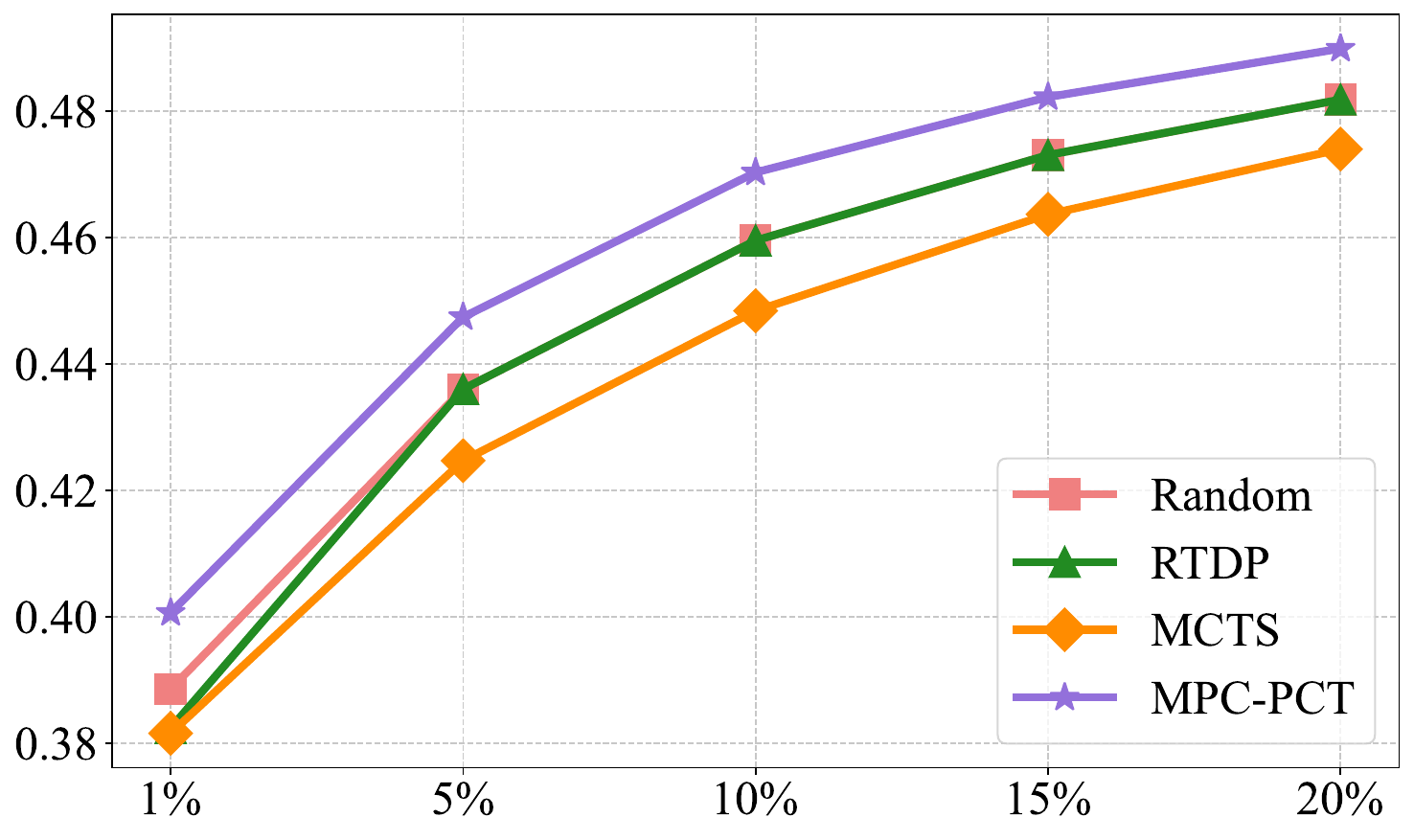}
}\hspace{\fourfigcol} 
\vspace{-1.25em}
\caption{Performance comparison under different levels.}
\vspace{-2em}
\label{fig: robustness}
\end{figure}
To evaluate the robustness of \textit{MPC-PCT}, we analyze its performance on subsets of bins ranked by final space utilization, specifically the top and bottom \{1\%, 5\%, 10\%, 15\%, 20\%\}. Figure~\ref{fig: robustness} plots the space utilization for each performance rank. We observe that \textit{MPC-PCT} (purple line) consistently achieves the highest space utilization across all levels, demonstrating its robustness. Among the baselines, \textit{RTDP} (green line) outperforms the \textit{Random} policy (pink line) in the median percentiles but underperforms at the extremes. This suggests that while an $\epsilon$-greedy exploration strategy may be effective for typical cases, a pure exploration approach can be more beneficial in the most extreme (easiest or hardest) scenarios.

\subsubsection{Analysis of scalability.}
\begin{figure}[t]
\centering
\subfigure[Lookahead Number $N$]{
\includegraphics[width=0.23\textwidth]{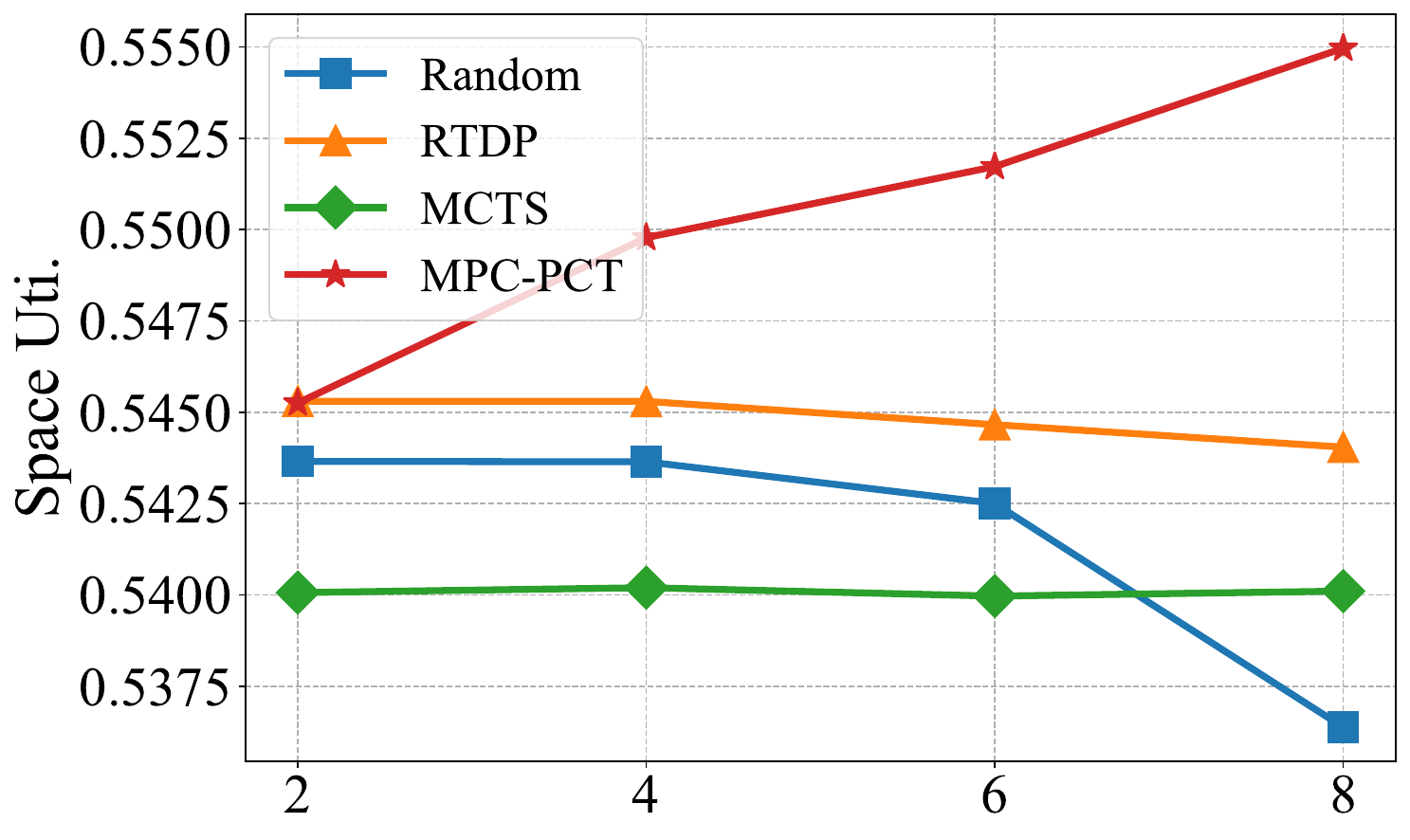}
}\hspace{\fourfigcol}
\subfigure[Lookahead Number $N$]{
\includegraphics[width=0.23\textwidth]{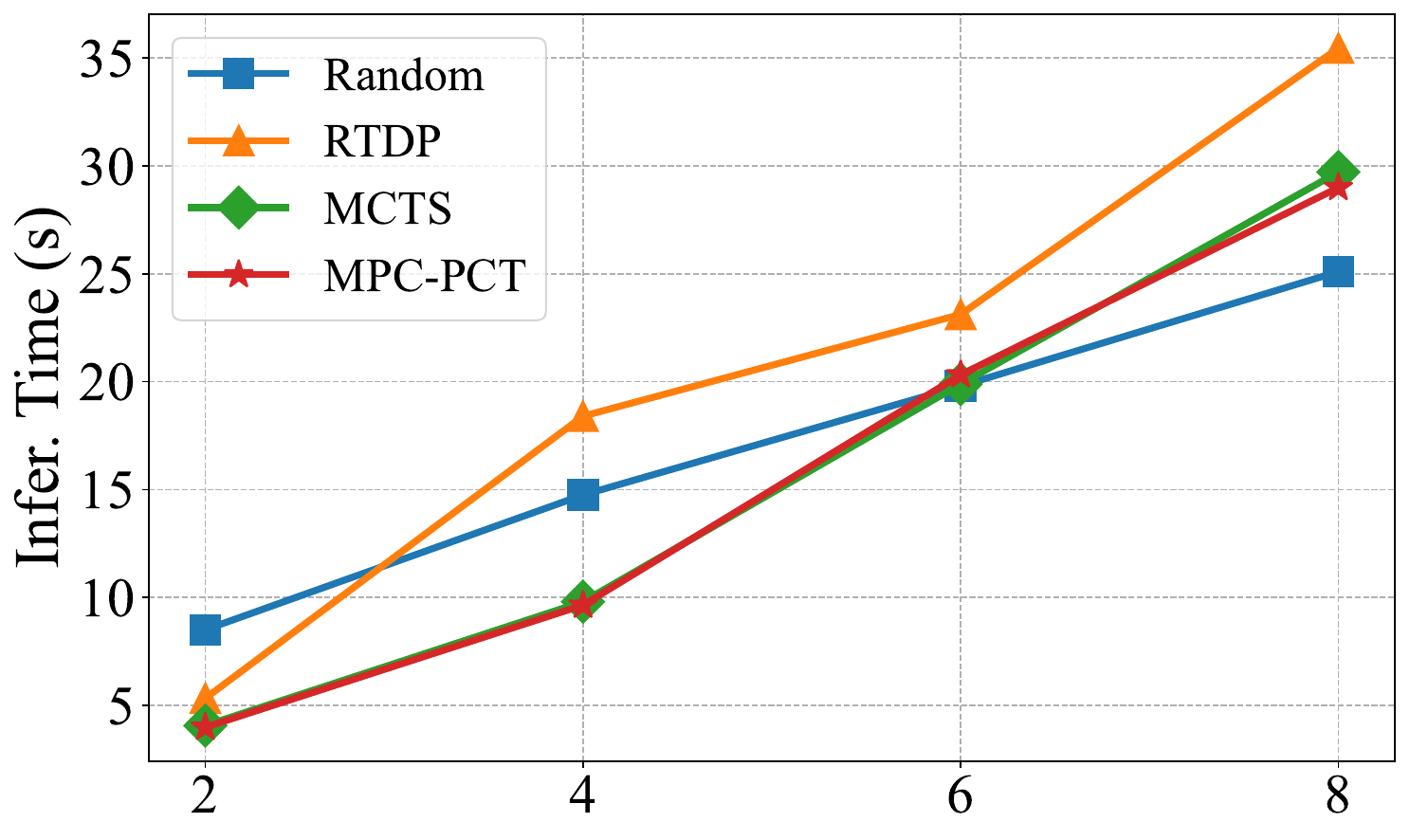}
}\hspace{\fourfigcol} 
\vspace{-1.2em}
\caption{Scalability}
\vspace{-1em}
\label{fig: scalability}
\end{figure}

To analyze the scalability of \textit{MPC-PCT} and other search techniques, we conduct experiments by varying the lookahead horizon $N$. The results are presented in Figure~\ref{fig: scalability}. We observe that only the performance of \textit{MPC-PCT} improves consistently as $N$ increases. Notably, the performance of the pure random exploration strategy (\textit{Random}, blue line) degrades dramatically with a larger $N$, eventually becoming worse than the baseline \textit{PCT} (which uses no lookahead). Furthermore, the performance of \textit{MCTS} and \textit{RTDP} remains largely unchanged as $N$ increases. This suggests that a proper balance between exploration and exploitation is crucial to effectively leverage a larger lookahead horizon. Additionally, the inference time of \textit{MPC-PCT} scales nearly linearly with $N$, which is consistent with its theoretical computational complexity of $O(nN)$.

\subsubsection{Effect of number of trials $n$.}
\begin{figure}[htbp]
\centering
\subfigure[Performance]{
\includegraphics[width=0.23\textwidth]{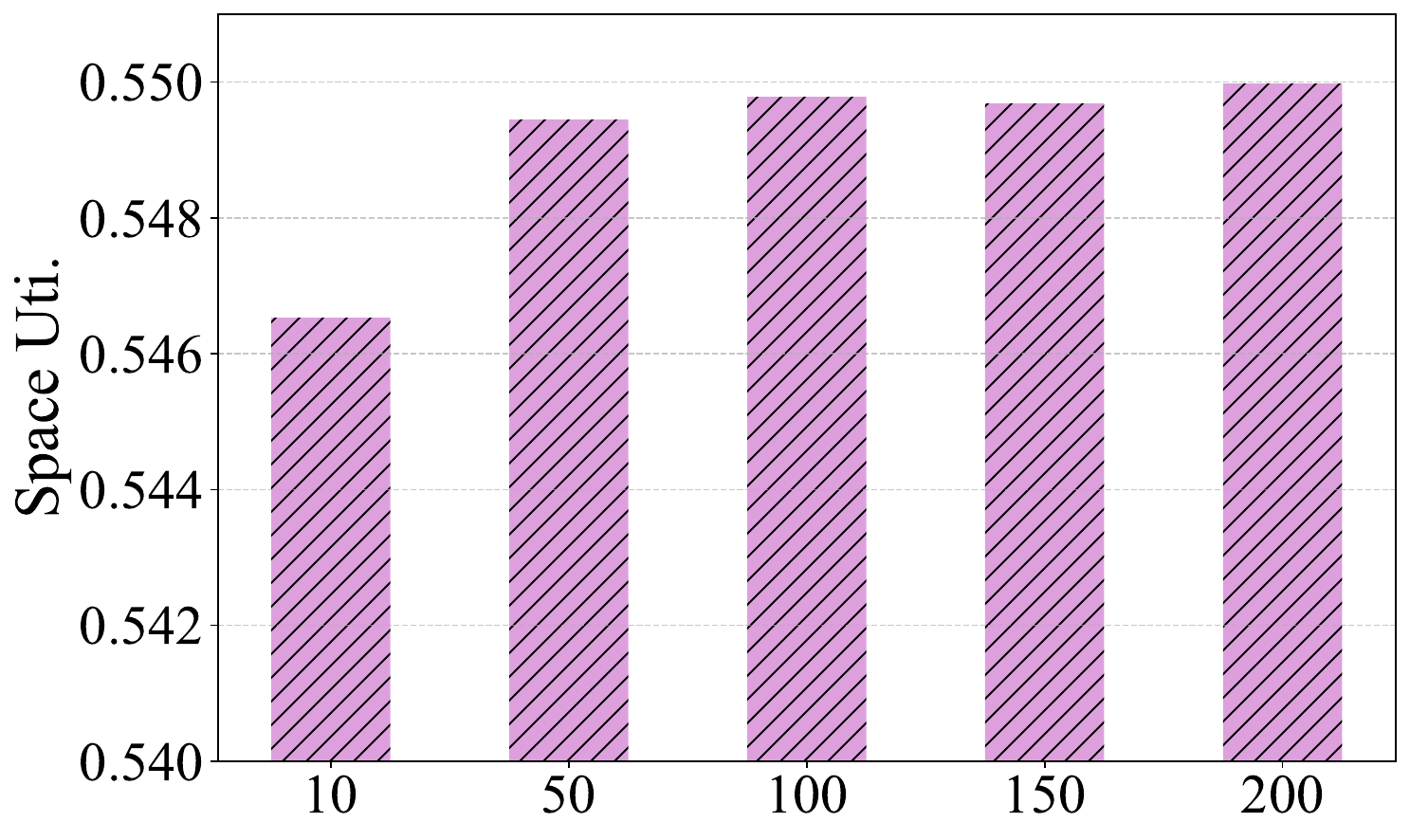}
}\hspace{\fourfigcol}
\subfigure[Efficiency]{
\includegraphics[width=0.23\textwidth]{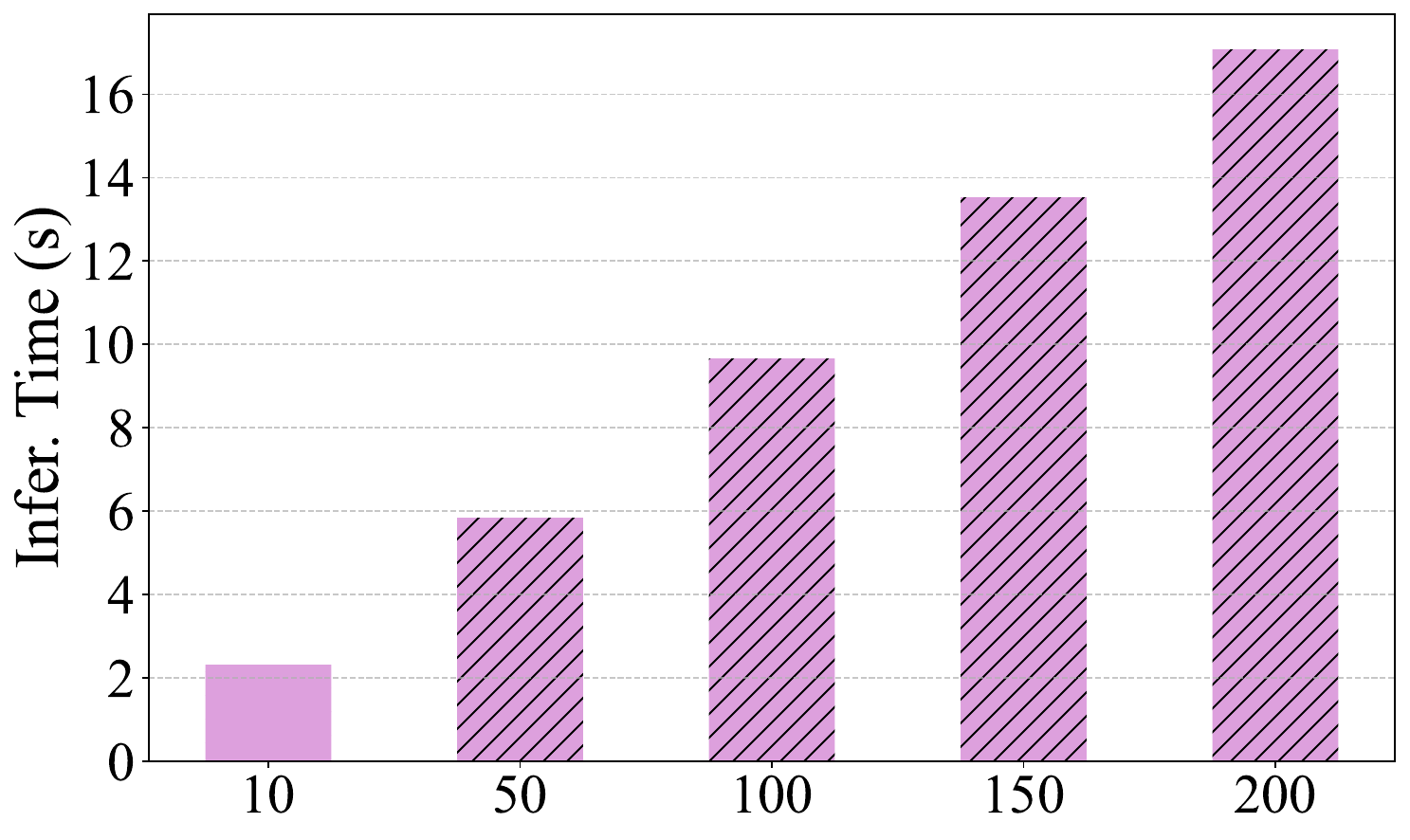}
}\hspace{\fourfigcol} 
\vspace{-1.25em}
\caption{The number of trials $n$ sensitivity.}
\vspace{-1.25em}
\label{fig: effect_n}
\end{figure}

Figure~\ref{fig: effect_n} shows the Space Utilization and Inference Time of \textit{MPC-PCT} as a function of $n$ (the total number of trials), while the lookahead horizon $N$ is held constant. We observe two key trends: (i) as $n$ increases, the Space Utilization improves and eventually plateaus; and (ii) the Inference Time increases linearly with $n$. Therefore, in practice, a trade-off can be made to achieve a sufficiently effective policy within a limited time budget. For instance, when $N=4$, setting $n=100$ yields near-optimal results in approximately 10 seconds.

\subsection{Case Study}
\begin{figure}[htbp]
  \centering
  \includegraphics[width=0.85\linewidth]{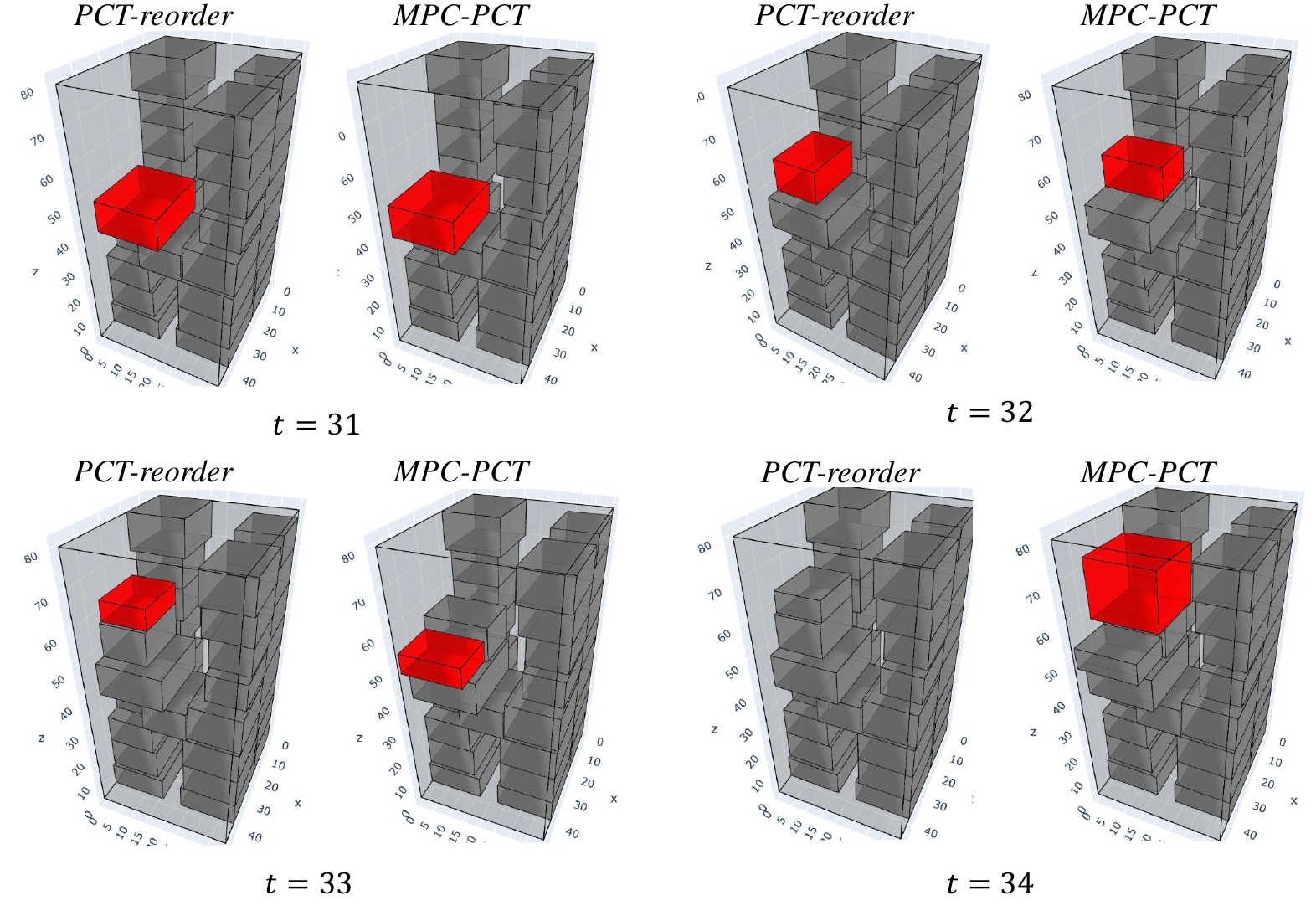}
  \vspace{-.75em}
  \caption{Case study on \textit{PCT-reorder} and \textit{MPC-PCT}. The red parcel is the current parcel to be packed at step $t$. If there is no red parcel means the bin is full.}
  \label{fig: case_study}
  \vspace{-1.25em}
\end{figure}
We visualize the bin packing results for \textit{PCT-reorder} and \textit{MPC-PCT} on an identical sequence of parcels. For a clear and fair comparison, we use the baseline \textit{PCT} policy to pack all parcels before step $t=31$, ensuring both algorithms begin with the same bin state. As shown in Figure~\ref{fig: case_study}, the incoming sequence from $t=31$ to $t=34$ consists of two extremely large parcels and two normally-sized ones. \textit{PCT-reorder} fails to pack the final parcel at $t=34$ because it still fundamentally relies on the learned \textit{PCT} policy and cannot sufficiently alter its behavior for this difficult sequence. In contrast, \textit{MPC-PCT} successfully packs the final parcel by adjusting its policy in advance; it rotates the parcels at $t=32$ and $t=33$ to create the necessary space for the item at $t=34$. This demonstrates our framework's ability to solve complex cases that simpler lookahead methods like \textit{PCT-reorder} cannot.

\section{Online Deployment}
\label{sec: online_deployment}
To validate the practical utility and real-world performance of our framework, we partnered with JD Logistics to deploy our framework in their intelligent bin packing system (Appendix~\ref{sec: deployment_system}).

\begin{figure}[htbp]
  \centering
  \includegraphics[width=0.85    \linewidth]{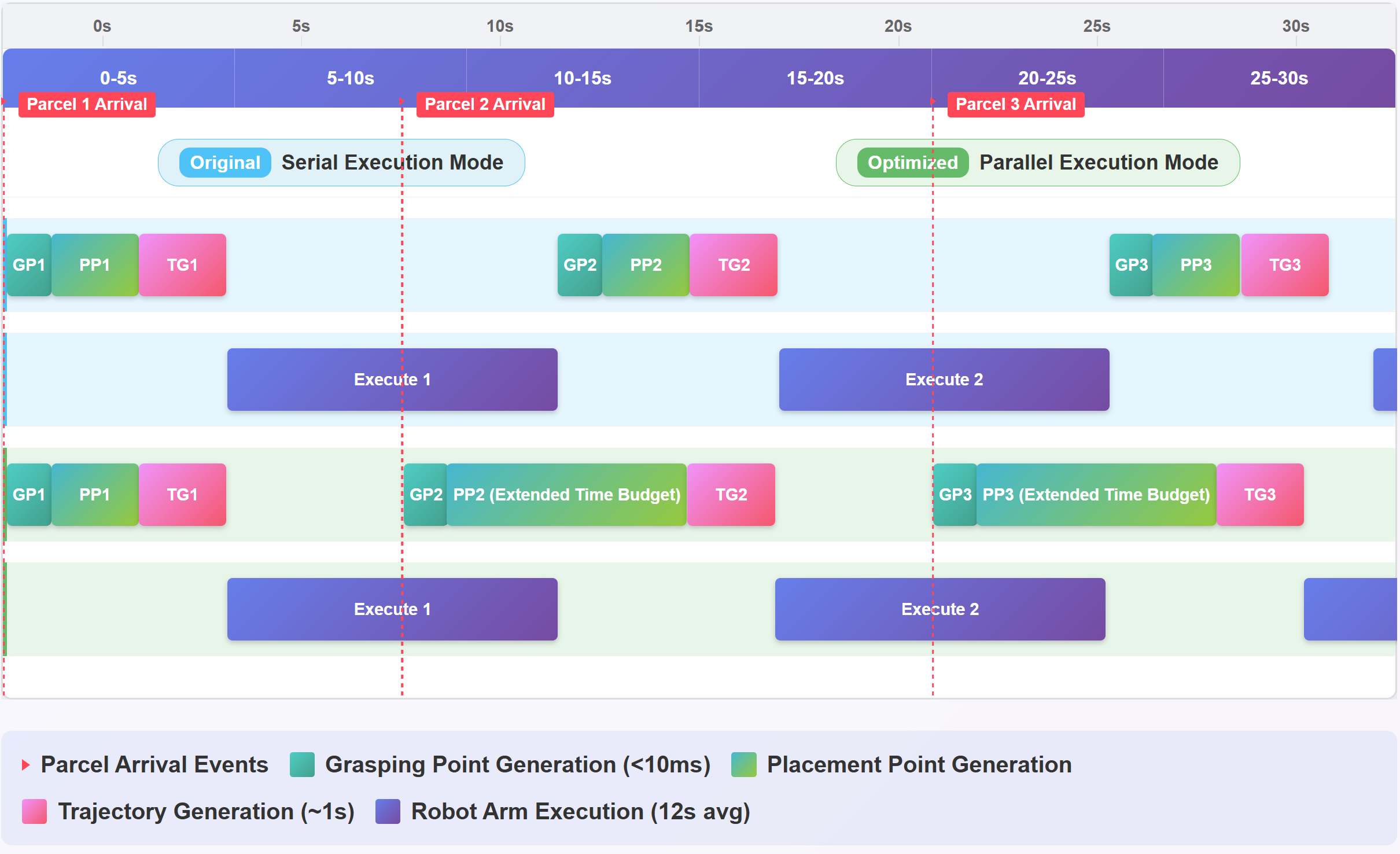}
  \vspace{-.75em}
  \caption{Engineering adjustment for system efficiency. Time interval for two consecutive parcels' arrival can be controlled to 15 seconds.}
  \label{fig: engineer_modification}
  \vspace{-2em}
\end{figure}

\newcommand{\subfigcol}{-2mm}
\begin{figure*}[t]
\begin{minipage}[c]{0.9\linewidth}
\centering
\subfigure{
\includegraphics[width=0.12\linewidth]{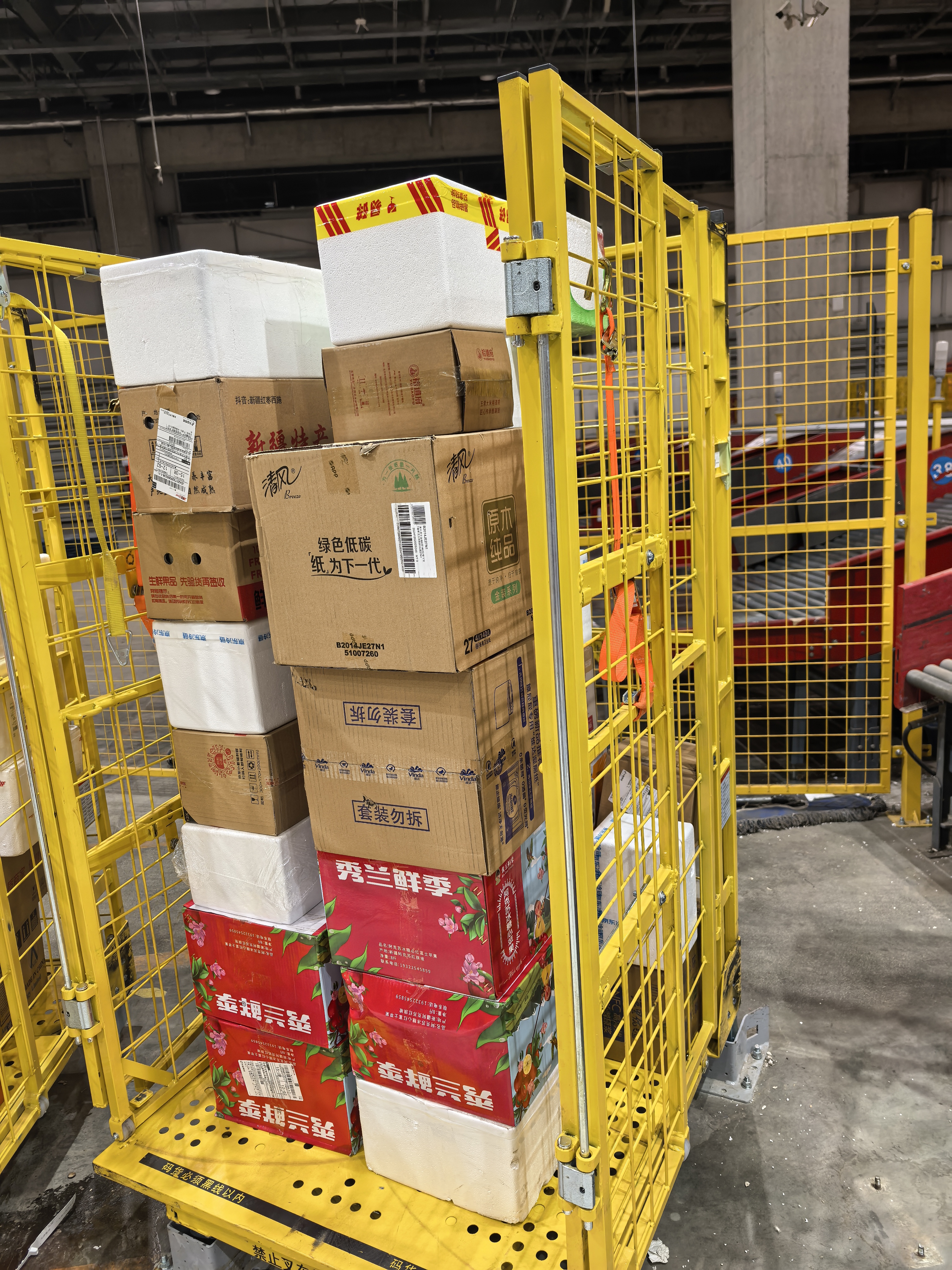}
    \label{vis_heatmap}
}\hspace{\subfigcol}
\subfigure{
\includegraphics[width=0.12\linewidth]{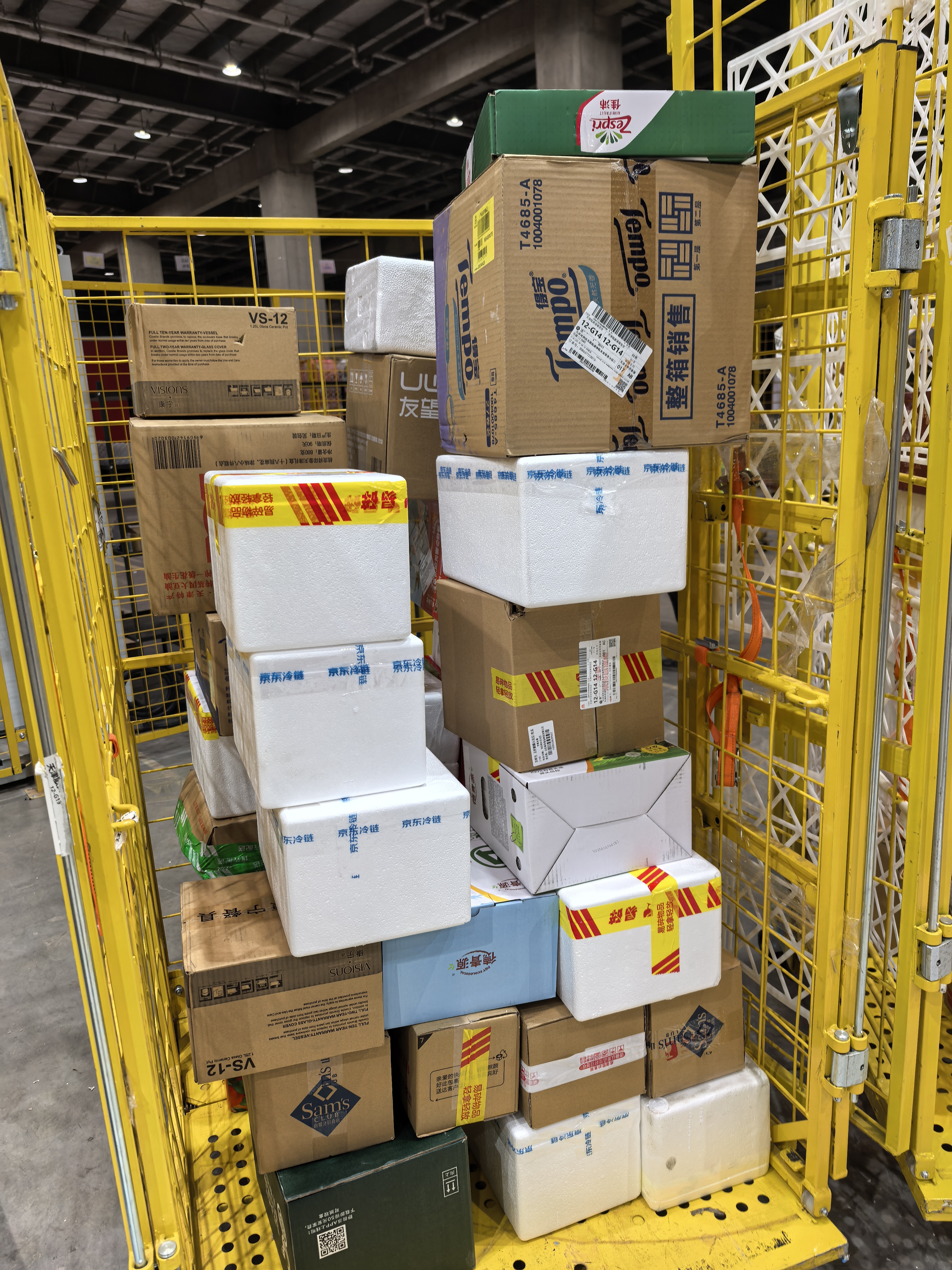}
    \label{vis_grid}
}\hspace{\subfigcol}
\subfigure{
\includegraphics[width=0.12\linewidth]{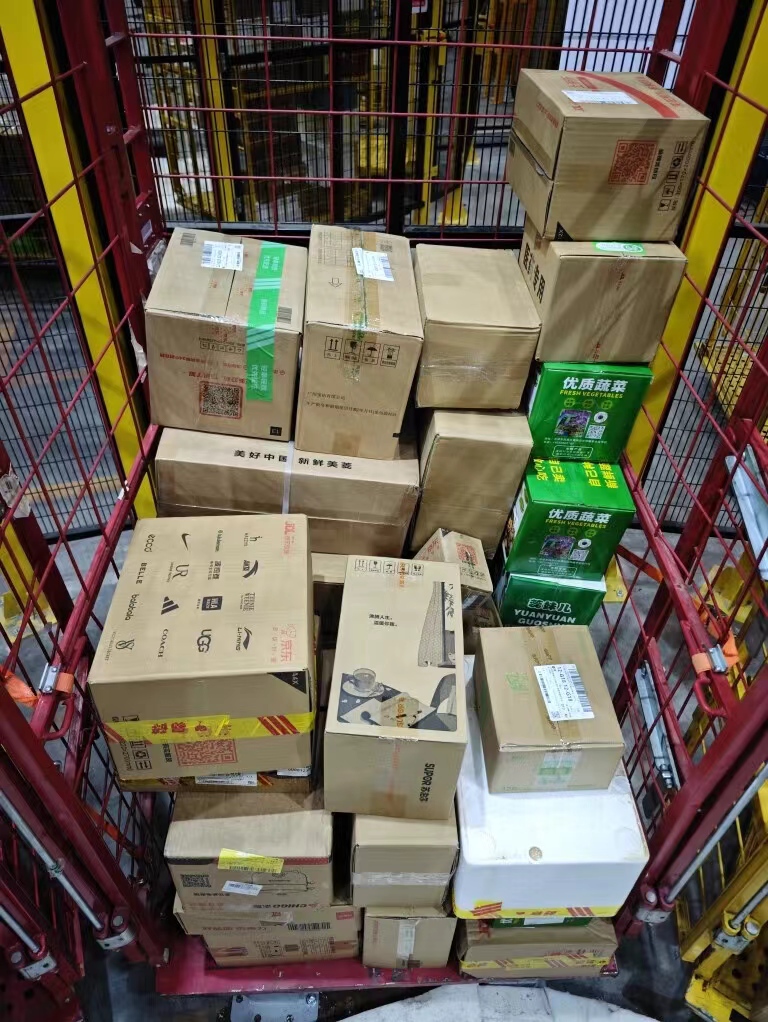}
    \label{vis_hexagon}
}\hspace{\subfigcol}
\subfigure{
\includegraphics[width=0.12\linewidth]{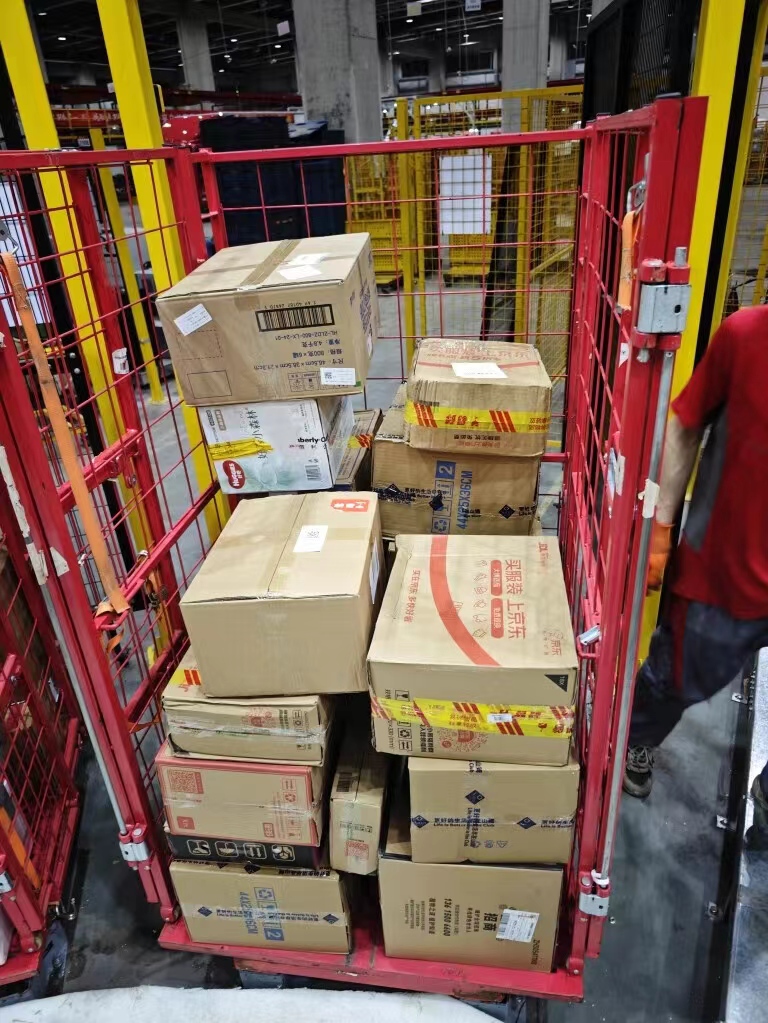}
    \label{vis_dbscan}
}\hspace{\subfigcol}
\subfigure{
\includegraphics[width=0.12\linewidth]{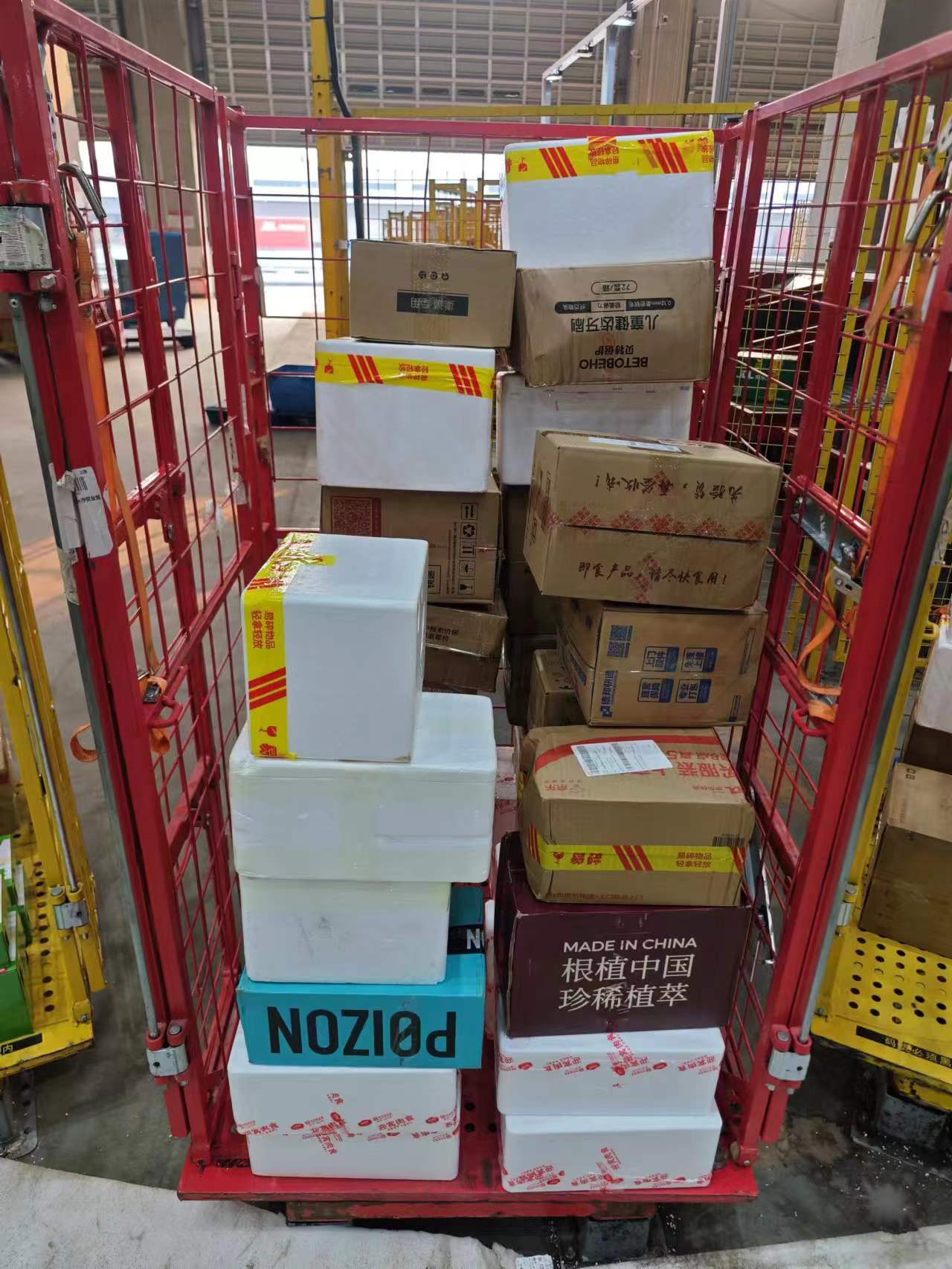}
    \label{vis_bsc}
}\hspace{\subfigcol}
\subfigure{
\includegraphics[width=0.12\linewidth]{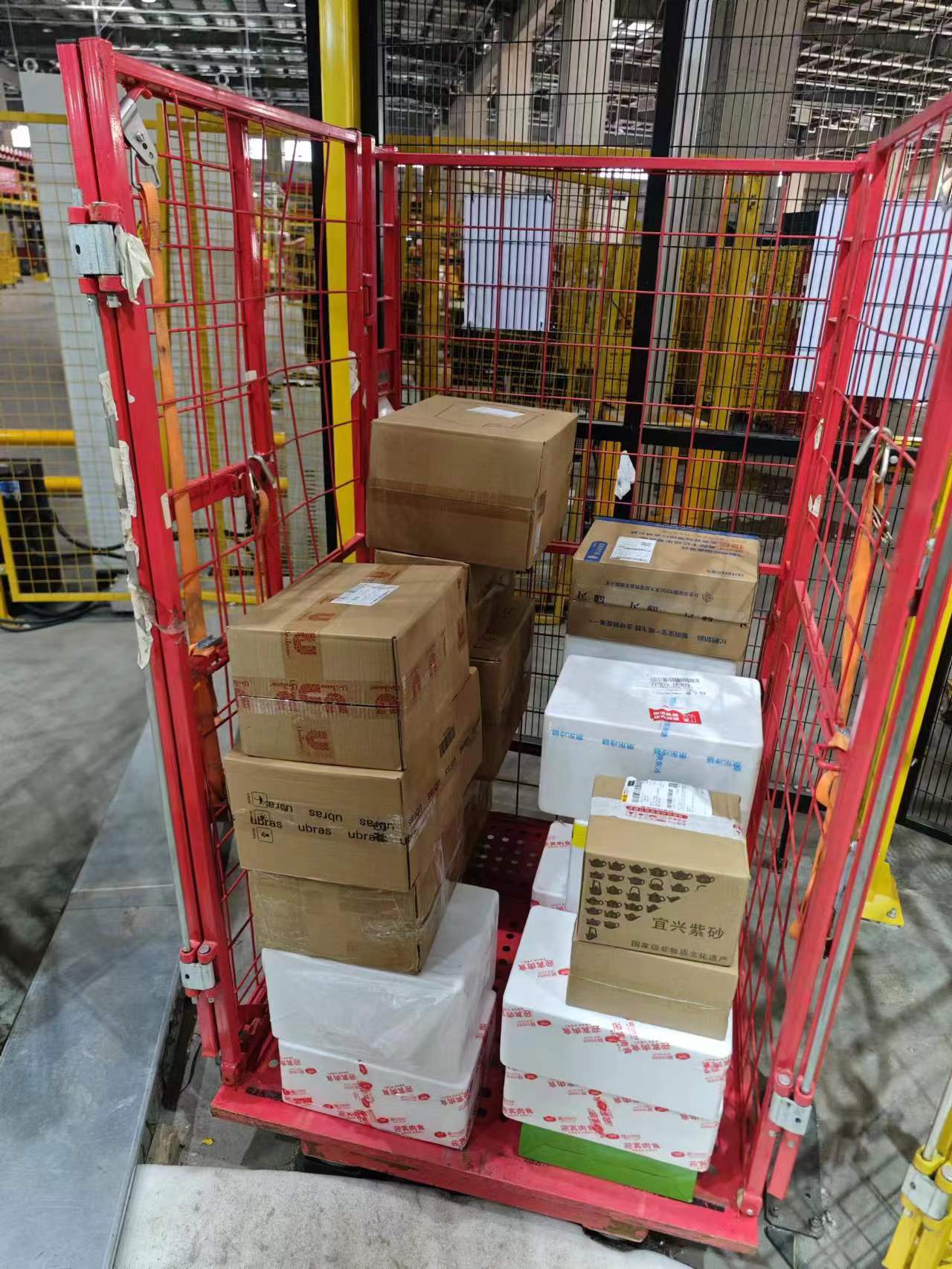}
    \label{vis_gcsc}
}\hspace{\subfigcol}
\subfigure{
\includegraphics[width=0.12\linewidth]{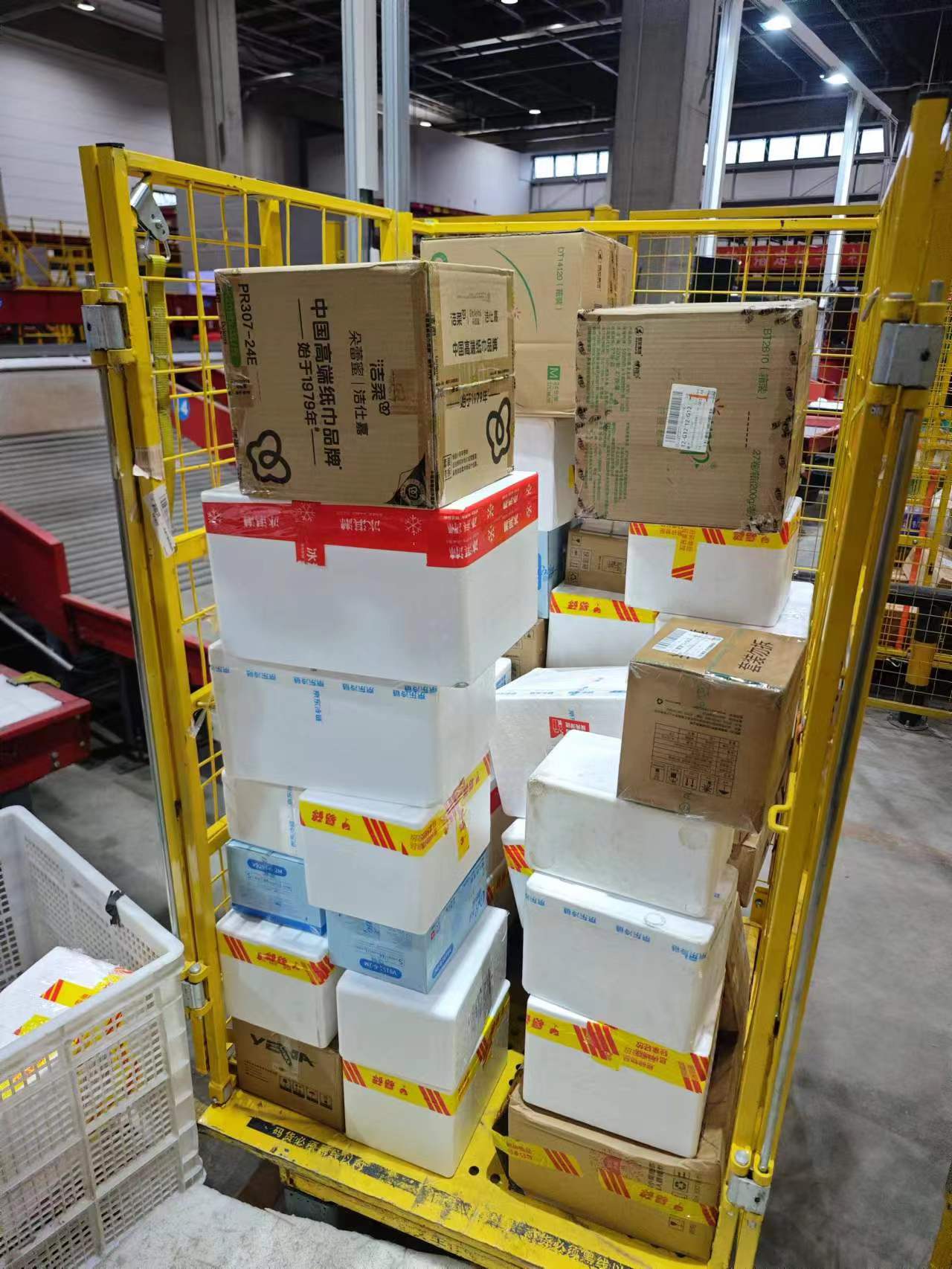}
    \label{vis_localsearch}
    }\hspace{\subfigcol}
\subfigure{
\includegraphics[width=0.12\linewidth]{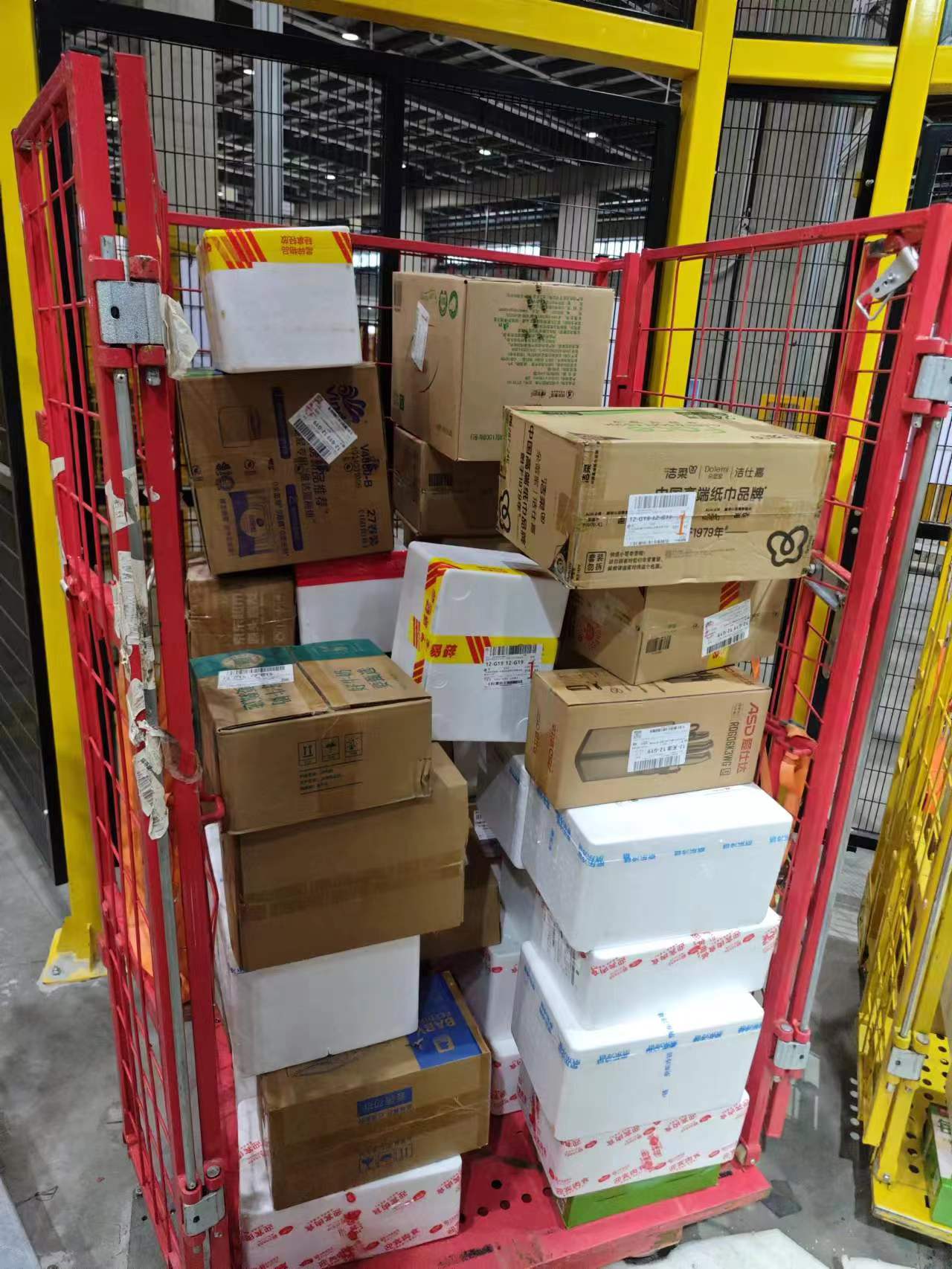}
    \label{vis_localsearch}
}\hspace{\subfigcol}
\vspace{-1.5em}
\caption{Real-world cases for our intelligent 3D-BP system with robotic arms.}
\label{fig: real_world_cases}
\end{minipage}
\vspace{-1.2em}
\end{figure*}
\subsection{Engineering Adjustment}

To deploy \textit{MPC-3D-BP} framework in the online robotic arm bin packing system, we first combine the well-trained production baseline model into our framework. 
Besides, we also try to optimize the efficiency at both the system level and the algorithm level. For system level, as shown in the original part of Figure~\ref{fig: engineer_modification}, the existing online system treats calculation (i.e., grasp point generation, placement point generation, and trajectory generation) and robotic arm execution as separate, sequential processes, leading to a tight time budget for calculation to prevent parcel accumulation. However, as shown in the optimized part of Figure~\ref{fig: engineer_modification}, robotic arm execution for the last parcel and calculation for the current parcel can run in parallel. This optimization can save at most about 10 seconds (i.e., extended time budget in Figure~\ref{fig: engineer_modification}) for placement point generation (i.e., Online 3D-BP algorithm), where $T_p =\min(T_a, T_e) - T_t-T_g\approx 12s-1s-10ms\approx11s$,$T_a$ for average time interval for consecutive parcels' arrival, $T_e$ for average time for robotic arm execution, $T_t$ for average time for trajectory generation, $T_g$ for average time for grasp point generation. For the algorithm level, we change the terminal condition for our algorithm from the number of trials to running time. In addition, since the current lookahead queue and the next lookahead queue have $N-1$ overlapped parcels, we cache the best child and the corresponding subtree of the root node to initialize the MCTS tree when the next parcel arrives.

\begin{figure}[htbp]
  \centering
  \includegraphics[width=0.85\linewidth]{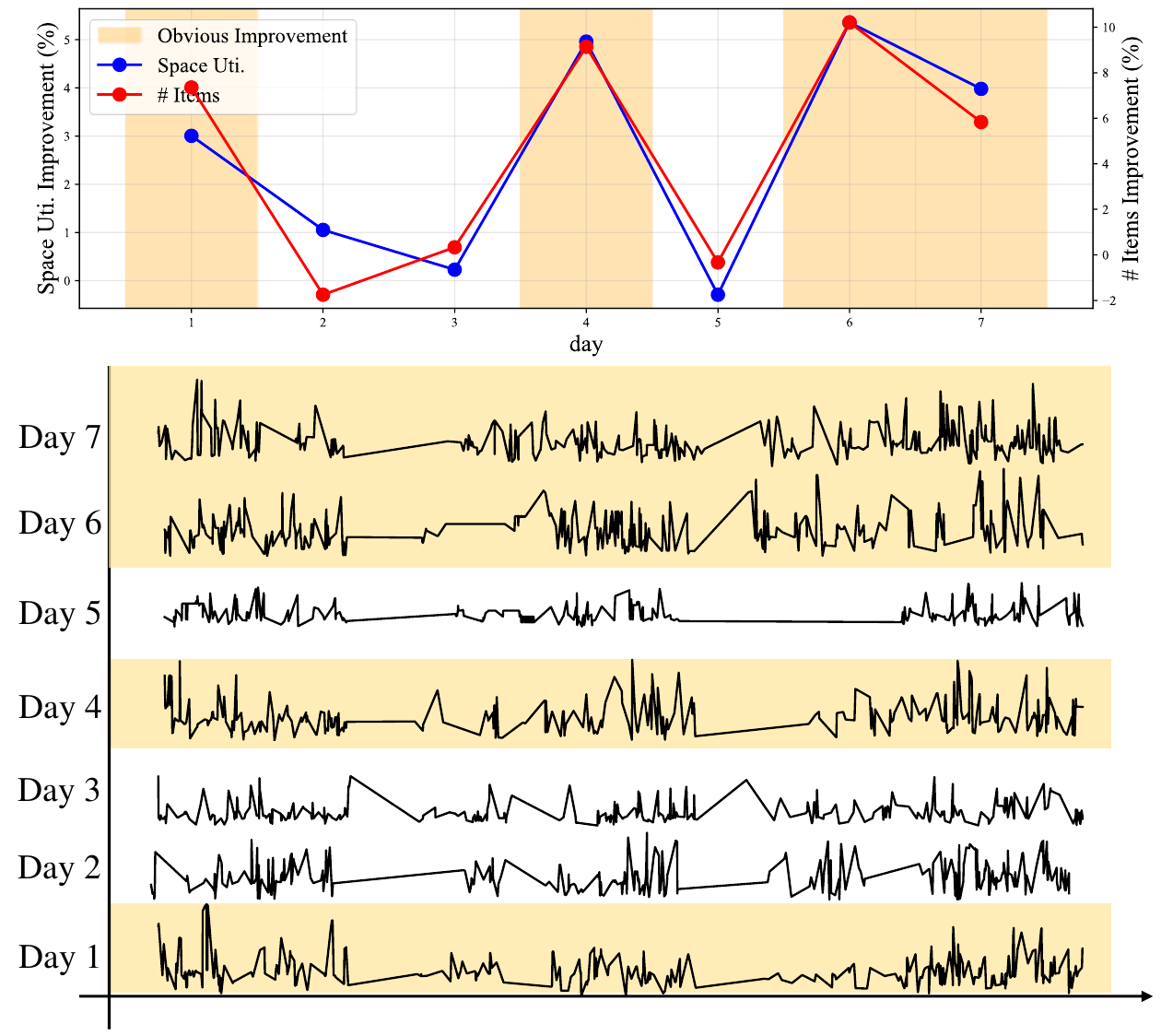}
  \vspace{-1em}
  \caption{Online deployment results for 7 days.}
  \label{fig: online_result}
  \vspace{-1.8em}
\end{figure}

\subsection{A/B Test Results}
We conduct online A/B tests on four parallel robotic arm lines, two of them equipped with the online 3D-BP DRL model (production baseline), and another two of them equipped with our proposed method. The results of the week-long online A/B test are presented in Figure~\ref{fig: online_result}. Our method achieves an average of 2.6\% higher space utilization and 4.4\% higher number of packed parcels compared to the production baseline. 

To understand why our model performs better on certain days (1, 4, 6, and 7, highlighted in orange), we analyze the corresponding parcel volume streams. The bottom of Figure~\ref{fig: online_result} plots the volume of the parcel (i.e., $l*w*h$) over a week, revealing distinct batches arriving throughout each day (e.g., morning, afternoon). The key observation is that the days with the largest performance gains coincide with the most volatile parcel streams (i.e., about $966\ \text{cm}^3$ alue of standard deviation\footnote{We calculate the standard deviation for 30 parcels as a window and average the standard deviation for all windows to obtain the final result}), whereas days with minimal improvement have more stable and consistent volumes (i.e., about $705\ \text{cm}^3$  standard deviation). Generally, we apply our methods in 3 sorting centers with 20 robotic arms. Since 2025-07 till now, we have already processed over 100000 parcels and packed over 2000 bins. Example real-world bin packing cases are shown in Figure~\ref{fig: real_world_cases}.

\section{Related Work}

Studies on the online 3D-BP problem generally fall into two categories: heuristic-based and DRL-based methods. Heuristic approaches codify human-designed strategies into rules, such as maximizing available space~\cite{ali2022line,wu2023machine}. Some offline heuristics can be directly applied in online settings~\cite{ali2022line}. Representative methods include OnlineBPH~\cite{ha2017online}, which introduces the Empty Maximal Spaces (EMS) heuristic prioritizing large empty regions; WallE~\cite{verma2020generalized}, which incorporates wall-building and layer-building strategies. In contrast, DRL-based methods learn placement policies through interaction with simulated environments. PackMan~\cite{verma2020generalized} was among the first to apply DRL in online 3D-BP. \citet{zhao2021online} introduced a feasibility mask to model the task as a constrained MDP, while \citet{zhao2021learning} proposed a Packing Configuration Tree (PCT) to guide DRL in continuous action spaces. AR2L~\cite{pan2023adjustable} designed a permutation-based attacker to maintain DRL performance under nominal and adversarial scenarios. GOPT~\cite{xiong2024gopt} employed a transformer to better capture spatial relations between parcels and the bin. Additionally, several studies enhance performance by enabling repacking actions inspired by human strategies~\cite{puche2022online,tsang2025deep,song2023towards,yang2023heuristics}. Lookahead information is pivotal in many sequential decision-making problems, such as Model Predictive Control (MPC) and Markov Decision Processes (MDPs)~\cite{merlis2024value,kouvaritakis2016model,merlis2024reinforcement}. In MDPs, numerous algorithms combine planning and RL to leverage lookahead, often refining base RL policy during training via search or optimization~\cite{silver2017mastering,efroni2019combine,efroni2020online,moerland2020think}. Alternatively, MPC-based methods solve a planning problem online at each step to achieve improved decision quality~\cite{hansen2022temporal}.

\section{Conclusion}
In this paper, we present \textit{MPC-3D-BP}, an online optimization framework based on Monte Carlo Tree Search for the online 3D-BP with lookahead parcels problem, designed to address the short-term distributional shifts common in real-world scenarios. To effectively balance exploration and exploitation, we propose a novel exploration strategy that is aware of potentially unfamiliar parcels in the lookahead queue. Additionally, we introduce a step-wise penalty to the reward function that penalizes wasted space, thereby accounting for the long-term impact of each placement. We conduct extensive experiments on datasets from both realistic and virtual settings. The results demonstrate that \textit{MPC-3D-BP} outperforms baselines that lack lookahead capabilities, and does so consistently when integrated with different underlying DRL models. Furthermore, we compare our framework against alternative problem formulations and MPC solution techniques, showing that our method achieves superior performance with an inference overhead that remains acceptable for online applications. Moreover, a week-long online A/B test on the live robotic packing system at JD Logistics confirms that \textit{MPC-3D-BP} outperforms the production baseline, verifying its real-world effectiveness. As one of the first studies on online 3D-BP with lookahead parcels, we hope our research can benefit the development of intelligent packing systems and offer insights for other similar problems involving finite lookahead horizons.

\bibliographystyle{ACM-Reference-Format}
\bibliography{ref}

\appendix

\section{Proof Details}
\label{sec: proof_details}
To quantify the suboptimality of the conventional policy $\bar{\pi}$ in a non-stationary environment, we will derive a bound on the performance gap. This bound will show that the loss in performance is directly related to the statistical distance between the long-term average distribution $D$ and the current short-term distribution $D_x$. We begin with a key lemma that bounds the change in the value of any fixed policy when the underlying environment distribution shifts from $D$ to $D_x$.

\noindent\textbf{Lemma 1.} Let $\pi$ be any fixed policy. Let $V^\pi_D(s)$ and $V^\pi_{D_x}(s)$ be the state-value functions of executing $\pi$ in environments governed by distributions $D$ and $D_x$, respectively. Assuming rewards are bounded in $\left[0,R_{max}\right]$ (in 3D-BP scenario, rewards corresponding to space utilization are in $[0,1)$), the maximum difference between these value functions is bounded as follows:
\begin{equation}
    \max\limits_s\lvert V_D^\pi(s) - V_{D_x}^\pi(s) \rvert \le \frac{\gamma R_{max}}{(1-\gamma)^2}\cdot D_{TV}(D,D_x)
\end{equation}
where $D_{TV}(D,D_x)$ is the Total Variation distance between the two distributions.

\noindent\textbf{Proof.}

\noindent Let $\delta(s)=V^\pi_D(s)-V^\pi_{D_x}(s)$. By the Bellman evaluation equation for a fixed policy $\pi$:
\begin{equation}
    \begin{split}
\delta(s)=\left(R(s,\pi(s))+\gamma\mathbb{E}_{d'\sim D}\left[ V_D^\pi(s') \right]\right)-\\ \left(R(s,\pi(s))+\gamma\mathbb{E}_{d'\sim D}\left[ V_{D_x}^\pi(s') \right]\right)
    \end{split}
\end{equation}

\begin{equation}
\delta(s)=\gamma\left(\mathbb{E}_{d'\sim D}\left[ V_D^\pi(s') \right]- \mathbb{E}_{d'\sim D}\left[ V_{D_x}^\pi(s') \right]\right)
\end{equation}
We add and subtract the term $\gamma \mathbb{E}_{d'\sim D}\left[V_D^\pi(s')\right]$ inside the absolute value:
\begin{multline}
\lvert \delta(s) \rvert
= \gamma \biggl\lvert \left( \mathbb{E}_{d'\sim D}\left[ V_D^\pi(s') \right] - \mathbb{E}_{d'\sim D_x}\left[ V_D^\pi(s') \right] \right) \\
+ \left( \mathbb{E}_{d'\sim D_x}\left[V_D^\pi(s')\right] - \mathbb{E}_{d'\sim D_x}\left[V_{D_x}^\pi(s')\right] \right) \biggr\rvert
\end{multline}
\begin{equation}
\begin{split}
    \left\lvert \delta(s) \right\rvert\le \gamma\left\lvert \mathbb{E}_{d'\sim D}\left[ V_D^\pi(s') \right] - \mathbb{E}_{d'\sim D_x}\left[ V_D^\pi(s') \right]\right\rvert + \\\gamma\left\lvert\mathbb{E}_{d'\sim D_x}\left[\delta\left(s'\right)\right]\right\rvert
\end{split}
\end{equation}
The value of any policy is bounded by $V_{max}=\frac{R_{max}}{1-\gamma}$. The difference between the expectations of a bounded function $f$ under two distributions is bounded by $2\vert\vert f \rvert\rvert_\infty D_{TV}$. Applying this to the first term:
\begin{equation}
    \left\lvert \delta(s) \right\rvert \le \gamma\left(2\cdot\frac{R_{max}}{1-\gamma}\cdot D_{TV}\left(D,D_x\right)\right) + \gamma \mathbb{E}_{d'\sim D_x}\left[\left\lvert \delta(s') \right\rvert \right]
\end{equation}
Let $\vert\vert \delta \rvert\rvert_\infty=\max_s \left\lvert \delta(s)\right\rvert$. Taking the maximum over all states $s$:
\begin{equation}
    \begin{aligned}
    \left\lvert\left\lvert \delta \right\rvert\right\rvert_\infty \le \frac{2\gamma R_{max}}{1-\gamma} D_{TV}\left(D,D_x\right) + \gamma \left\lvert\left\lvert \delta \right\rvert\right\rvert_\infty\\
    \left(1-\gamma\right)\left\lvert\left\lvert \delta \right\rvert\right\rvert_\infty\le \frac{2\gamma R_{max}}{1-\gamma} D_{TV}\left(D,D_x\right) 
    \end{aligned} 
\end{equation}
This yields the bound (a slightly different constant due to the bounding technique, but the principle holds):
\begin{equation}
    \left\lvert\left\lvert \delta \right\rvert\right\rvert_\infty\le \frac{2\gamma R_{max}}{(1-\gamma)^2} D_{TV}\left(D,D_x\right)
\end{equation}
This lemma formalizes that a small change in distribution leads to a proportionally small change in the value of any given policy. With this lemma, we can now prove our main theorem.

\noindent\textbf{Theorem 2 (Performance Gap Bound)}. The performance gap of the conventional policy $\bar{\pi}$ in an environment governed by $D_x$ is bounded by a function of the statistical distance $D_{TV}(D,D_x)$.

\noindent\textbf{Proof}.
The performance gap at a state $s$ is defined as the difference between the true optimal value in the $D_x$ environment and the value achieved by our conventional policy $\bar{\pi}$ in that same environment:
\begin{equation}
    \text{Gap}(s)=V^*_{D_x}(s)-V^{\bar{\pi}}_{D_x}(s)
\end{equation}
We know that the policy $\bar{\pi}$ is optimal for the stationary distribution $Q$. Therefore, its value function is the optimal one in that environment: $V^{\bar{\pi}}_D(s)=V^*_D(s)$.

We can rewrite the gap by adding and subtracting the value of the optimal policy for the stationary world, $V^*_D(s)$:
\begin{equation}
    \text{Gap}(s) = \left(V^*_{D_x}(s)-V^*_{D}(s)\right) + \left(V^*_D(s)-V^{\bar{\pi}}_{D_x}(s)\right)
\end{equation}
Substituting $V^*_D(s)=V^{\bar{\pi}}_D(s)$:
\begin{equation}
\begin{split}
        \text{Gap}(s) = 
\underbrace{\left(V^*_{D_x}(s) - V^*_{D}(s)\right)}_{\text{Term 1: Inherent Optimality Difference}} 
+ \\
\underbrace{\left(V_{D}^{\pi}(s) - V_{D_x}^{\pi}(s)\right)}_{\text{Term 2: Policy Performance Degradation}}
\end{split}
\end{equation}
This decomposition is insightful. It shows that the total performance gap comes from two sources:
\begin{enumerate}
    \item \textbf{Term 1}: The difference between the best possible performance in the new environment $V^*_{D_{x}}$ versus the old environment $V^*_D$. This reflects how the environment's change inherently affects the optimal achievable value.
    \item \textbf{Term 2}: The degradation in performance of our specific, fixed policy $\bar{\pi}$
\end{enumerate}
Using \textbf{Lemma 1}, we can directly bound Term 2:
\begin{equation}
    \left\lvert V_D^{\bar{\pi}}(s) - V_{D_x}^{\bar{\pi}}(s)\right\rvert \le \frac{2\gamma R_{max}}{(1-\gamma)^2} D_{TV}\left(D,D_x\right)
\end{equation}
A similar, though more involved, proof based on the properties of the Bellman optimality operator as a contraction mapping can show that Term 1 is also bounded by the distribution distance:
\begin{equation}
     \left\lvert V_{D_x}^*(s) - V_{D}^*(s)\right\rvert \le \frac{2\gamma R_{max}}{(1-\gamma)^2} D_{TV}\left(D,D_x\right)
\end{equation}
By applying the triangle inequality, the total performance gap is thus bounded:
\begin{equation}
    \lvert\text{Gap}(s)\rvert\le \lvert V_{D_x}^*(s) - V_{D}^*(s) \rvert + \lvert V_D^{\bar{\pi}}(s) - V_{D_x}^{\bar{\pi}}(s) \rvert \le C\cdot D_{TV}(D,D_x)
\end{equation}
where $C=\frac{4\gamma R_{max}}{(1-\gamma)^2}$is a constant.

Besides, the proof is already applicable to MDPs that can end in finite horizons where $\gamma$ is often set to 1. Although it may seem that when $\gamma = 1$, the constant $C$ goes to $\infty$, in fact, if the game can end in a maximum of $T$ finite steps, the value of any policy is still bounded by $T \cdot R_{\text{max}}$. Other proof details remain unchanged.

\section{Data Evidence and Additional Experiment Results}
\label{sec: illustrative_example_data_evidence}
\begin{figure}[htbp]
  \centering
  \includegraphics[width=\linewidth]{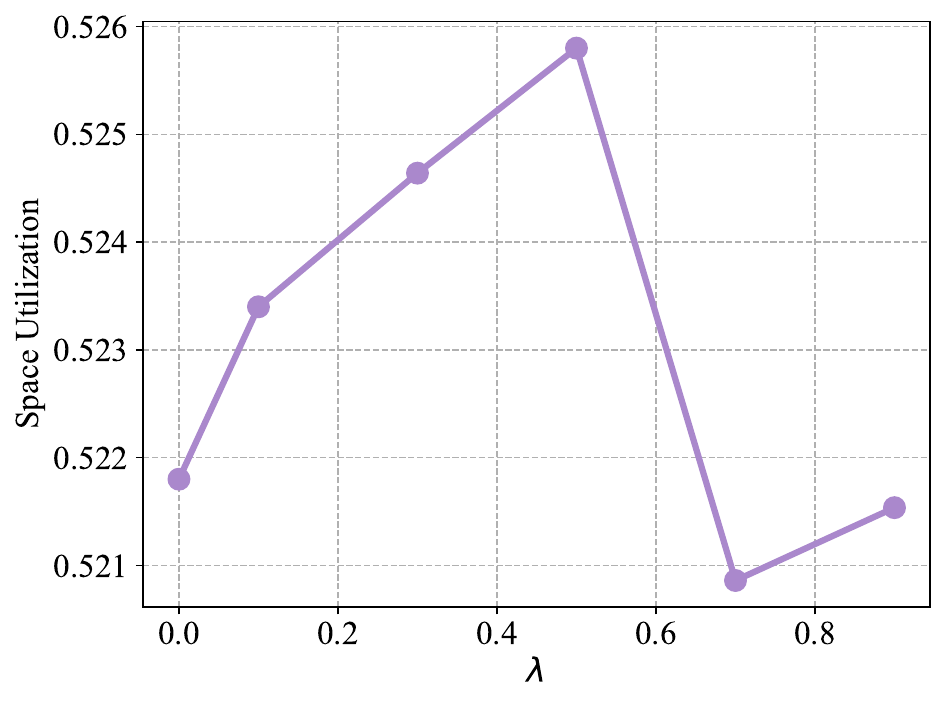}
  \caption{Parameters sensitivity study on waste space penalty $\lambda$ in the \textbf{Shift} scenario.}
  \label{fig: lambda}
\end{figure}


\begin{figure}[htbp]
  \centering
  \includegraphics[width=\linewidth]{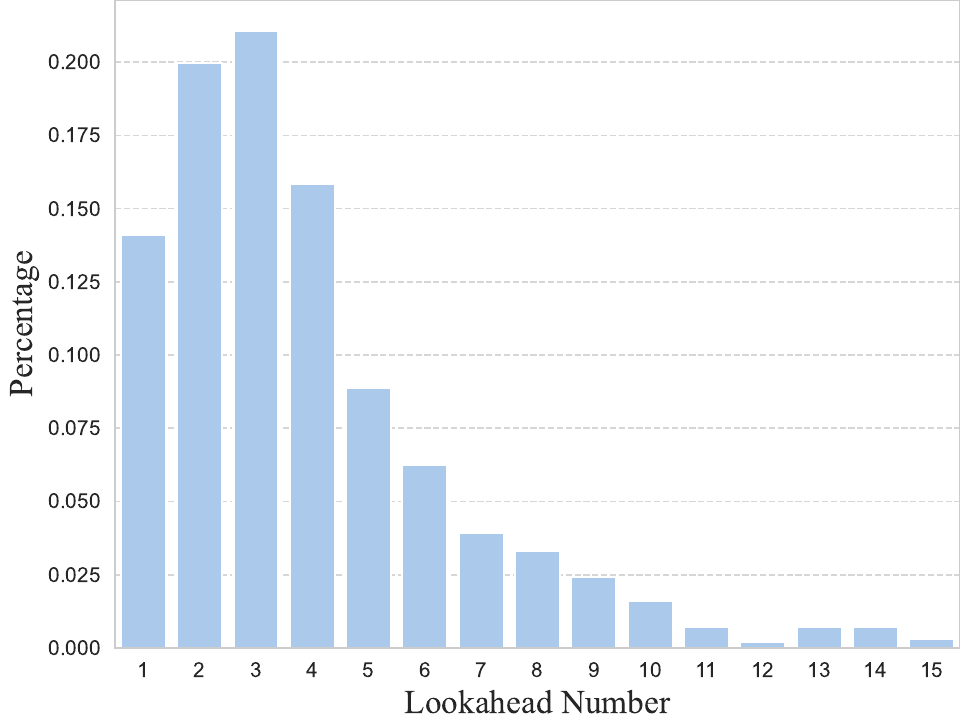}
  \caption{Common lookahead number $N$ in Cyber-Physical System.}
  \label{fig: lookahead}
\end{figure}

\begin{figure}[htbp]
  \centering
  \includegraphics[width=\linewidth]{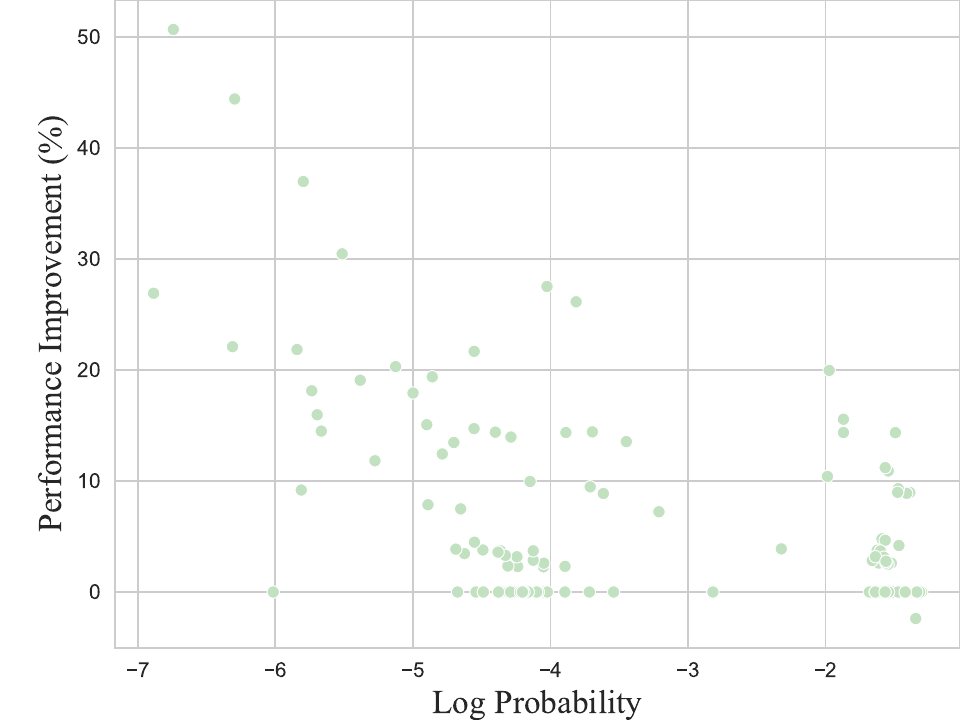}
  \caption{The log probability of online parcel batches, calculated relative to the offline training dataset, plotted against the corresponding performance gap between the \textit{BFS} and \textit{PCT} methods.}
  \label{fig: log_p_vs_Performance_gap}
\end{figure}

\section{Implementation Details}
\label{sec: implementation_details}
\subsection{Dataset Preprocess and Statistics}
The \textbf{offline training dataset} is mined from the historical robotic arm log in JD Logistics. We collect the length, width, and height of the parcels (detailed statistics in Table~\ref{table: statistics}), as well as their grasp points, to serve as training inputs. All data are obtained under consistent real-world production conditions. The inclusion of grasp points as input features during training is determined by the specific experimental settings.

The \textbf{lookahead parcels dataset} is also created by mining historical robotic arm logs from JD Logistics. The raw logs contain detailed records for each parcel, including its dimensions (length, width, and height), arrival time, and the corresponding robotic arm actions (e.g., grasp points and placement coordinates). We extract this essential information and then sort the records by their arrival times to reconstruct the original parcel stream. To generate the lookahead information for each record, we first set a lookahead horizon $N$. Then, for each placement decision, its lookahead queue is formed by taking the next $N-1$ parcels from the sorted stream. The total number of records for the lookahead parcels dataset is 882.

\begin{table}[htbp]
    \caption{Dataset Statistics of offline training dataset.}
    \resizebox{0.48\textwidth}{!}{\textcolor{blue}{}
    \begin{tabular}{ccccccc}
        \toprule
         \textbf{Time Span} & \textbf{\# Records} & \textbf{Length Range} & \textbf{Width Range} & \textbf{Height Range} & \textbf{Bin Size} \\ 
         \midrule
          2025.05-2025.06 & 5652 & $16\sim58$ & $12\sim40$ & $4\sim48$ & (94,80,160)\\
        \bottomrule
    \end{tabular}}
    \label{table: statistics}
\end{table}

\subsection{Baselines}
\label{sec: baselines}
Heuristic methods are important baselines in the Online 3D-BP problem.
\begin{itemize}[leftmargin=0.2cm]
    \item \textit{LSAH}~\cite{hu2017solving}: {According to~\citeauthor{hu2017solving}, a heuristic algorithm which goes over all empty maximal spaces and 2 orientations for the current item and chooses the empty maximal space and orientation that yields the least surface area when putting an item.}
    \item \textit{OnlineBPH}~\cite{ha2017online}: {According to~\citeauthor{ha2017online}, a heuristic algorithm which finds the first suitable placement within the allowed empty maximal spaces and orientations. All the empty maximal spaces are sorted with deep-bottom-left order.}
    \item \textit{MACS}~\cite{HuXCG0020}: {According to~\citeauthor{HuXCG0020}, a heuristic algorithm which calculates the best-scored position by the policy of maximizing remaining empty spaces. Candidate locations are generated from the four corners of every empty maximal space under two possible orientations.}
    \item \textit{DBL}~\cite{karabulut2004hybrid}: {According to~\citeauthor{karabulut2004hybrid}, a heuristic algorithm which selects positions based on the Deepest Bottom Left with Fill (DBLF) policy, which prioritizes the deepest available slot, followed by the lowest, and then the leftmost position.}
\end{itemize}
Besides, we implement two methods considering lookahead information:
\begin{itemize}[leftmargin=0.2cm]
    \item \textit{PCT-lookahead}: a DRL method that encodes lookahead parcels as part of states.
    \item \textit{PCT-reorder}: Following the work of~\citeauthor{zhao2021online}, we implement a reordering algorithm that conditions the current placement decision on upcoming parcels. This method virtually reorders the sequence to place lookahead items first, while adhering to a critical order-dependence constraint: an earlier-arriving parcel cannot occupy a space made available only by the placement of a later-arriving one.
\end{itemize}

In addition, we choose two DRL methods as our backbone, where \textit{PCT} is our main backbone and we also test our method's effectiveness on \textit{GOPT}.
\begin{itemize}[leftmargin=0.2cm]
    \item \textit{PCT} is a state-of-the-art DRL model for online 3D-BP problem. It, through its novel tree-based representation, cleverly combines heuristic rules and deep reinforcement learning using the Graph Attention Network (GAT)~\cite{yu2023practical} to solve the core challenge of an overly complex action space in the online 3D Bin Packing problem.
    \item \textit{GOPT} is a state-of-the-art DRL model for online 3D-BP problem. It, through its unique "Generator-Transformer" architecture, achieves high performance and strong generalization for the 3D bin packing task, addressing the problem where traditional DRL methods often struggle to adapt to different environmental configurations.
\end{itemize}

To validate the effectiveness of our adaptation, we also implement many alternative general techniques to solve the same MPC problem.

\begin{itemize}[leftmargin=0.2cm]
    \item \textit{BFS}~\cite{zhou2006breadth}: We implement a brute-force search that traverses every possible root-to-leaf path to identify and select the optimal one.
    \item \textit{Random}~\cite{pemantle1995critical}: We implement many random walks on the tree where, for each path traversal, the algorithm starts at the root and randomly selects a child node to visit at each step until a leaf node is reached. After performing a predefined number of searches, we select the optimal path found among all explored trajectories.
    \item \textit{RTDP}~\cite{barto1995learning}: According to~\citeauthor{barto1995learning}, we implement a real-time dynamic programming algorithm in the context of the online 3D-BP problem.
    \item \textit{MCTS}~\cite{silver2018general}: According to~\citeauthor{silver2018general}, we adapt the MCTS framework used in AlphaZero directly to the 3D-BP environment.
\end{itemize}
\subsection{Metrics}~\label{sec: metric}
The space utilization (Space Uti.) is calculated as follows:
\begin{equation}
    \text{space\_utilization} = \sum\limits_{i=1}^M\frac{l_i*w_i*h_i}{L*W*H}
\end{equation}
where $M$ is the number of parcels for the bin, $L,W,H$ are the length, width, and height for the bin, respectively, $l_i,w_i,h_i$ is the length, width, and height for the $i^{th}$ parcel, respectively.

The number of parcels (\# Items)  is directly defined by its literal meaning.

\subsection{Experiment Settings}
\noindent\textit{Training}. In the offline training stage, we train two models: \textit{PCT} and \textit{GOPT}. For \textit{PCT} training, we use the Empty Maximal Spaces (EMS) method to generate the action space and encode the state using a single graph attention layer. The hyperparameters for both deep models are fine-tuned via grid search. Key architectural parameters include an input embedding dimension of 64, a hidden layer dimension of 128, an internal node length of 120, and a leaf node length of 80. Additionally, the hyperparameters for the learning process are as follows: the learning rate is set to $1 \times 10^{-6}$, and the coefficients for both the actor and critic losses are set to 1.0. For \textit{GOPT} training, we use EMS as the action space and encode the state using three transformer layers. When the model selects an EMS as the output, the parcel is placed at the deepest, lowest, and leftmost feasible position within the selected EMS. Key architectural parameters include an input embedding dimension of 128, an EMS length of 80. Additionally, the hyperparameters for the learning process are as follows: the learning rate is set to $7 \times 10^{-5}$ at first with a linearly descending learning rate, and the coefficients for policy update and critic loss are respectively set to 0.96 and 0.5.
\begin{figure}[htbp]
  \centering
  \includegraphics[width=0.8\linewidth]{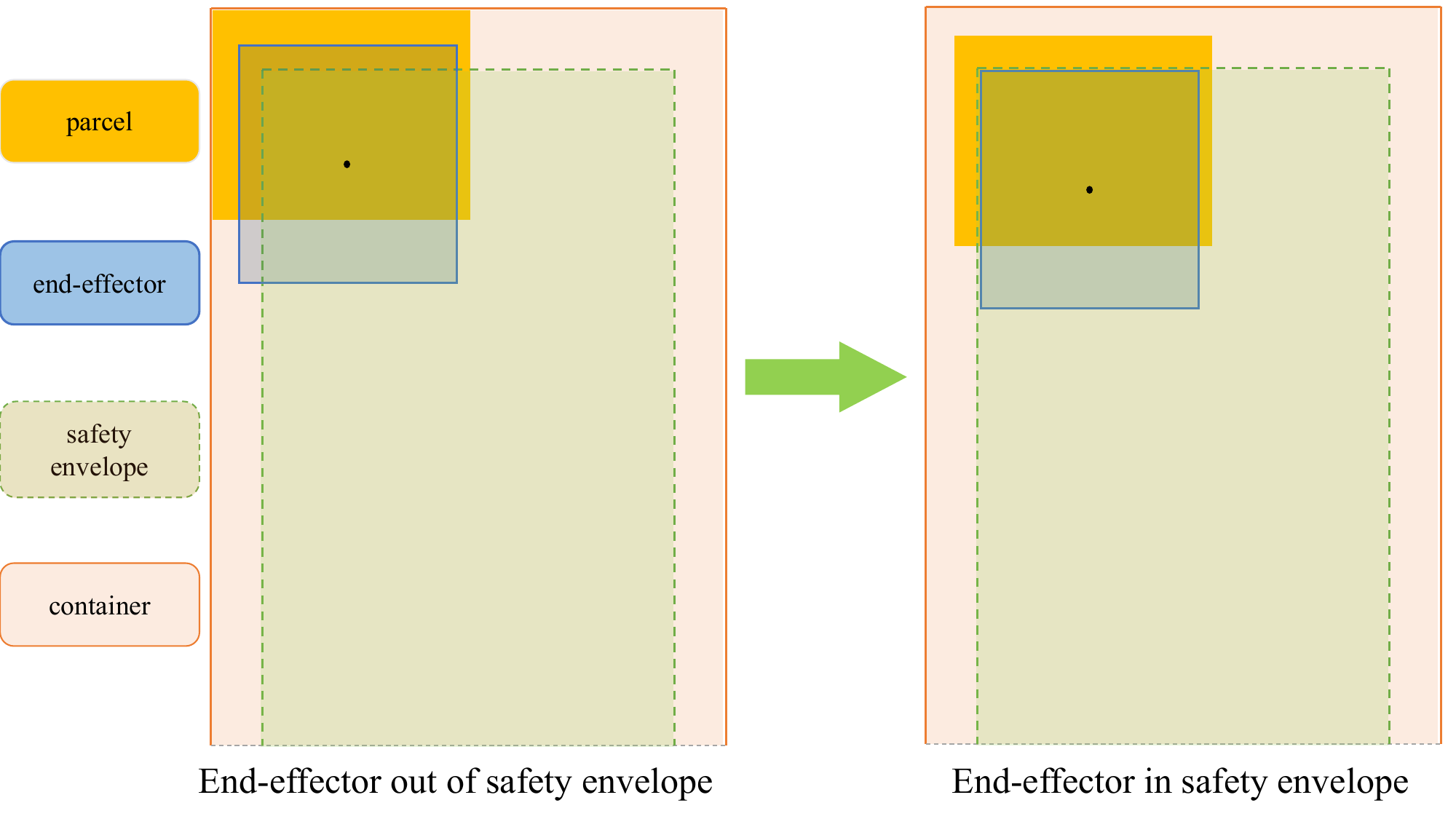}
  \caption{Top view of the position safety offset method.}
  \label{fig: grasp_point}
\end{figure}

\noindent\textit{Testing}. We test two settings of the same dataset, \textbf{Realistic} and \textbf{Virtual}.
\begin{itemize}[leftmargin=0.2cm]
    \item \textbf{Virtual}. To comprehensively evaluate the performance of all methods on the parcel stream dataset, we use a sliding-window technique to generate short-term test batches that simulate the online scenario. Specifically, we set the window size to 100, which corresponds to the maximum number of items historically packed into a single bin, and generate these batches using a stride of 1. To evaluate a given method, we run it on each batch and then calculate the final performance by averaging the metrics across all resulting packed bins. Besides, we follow \textit{Setting} 1 in \textit{PCT}~\cite{zhao2022learning}. Only two horizontal orientations are permitted, and the stability of the bin state is checked when the current parcel is placed.
    \item \textbf{Realistic}. With the same evaluation protocol of \textbf{Virtual} setting, we consider more practical constraints working with robotic arms. In the \textbf{Realistic} setting, the data includes the grasp point information of the robotic arm when picking up parcels, which is used to ensure operational safety in real-world scenarios. Given the constraints and limitations characteristic of real-world production environments, the robotic arm’s end-effector is at risk of colliding with the cage walls or previously placed parcels during item pick-and-place operations. To mitigate this risk, we define a safety zone—referred to as the safety envelope—within the cage, as illustrated in the figure~\ref{fig: grasp_point}, where the end-effector is only allowed to operate within the green area. Based on the upstream-provided grasp points, we calculate whether the current placement position would cause the end-effector to exceed the permissible range. Notably, when the parcel size is very small, the planned position may result in the end-effector exceeding the safety zone, rendering many placements infeasible. To address this issue, we designed a position safety offset method that computes the minimally offset placement position while still satisfying the safety requirements (Figure~\ref{fig: grasp_point}). This offsetting approach applies to all action generation schemes as well as heuristic methods. We apply this method in all realistic settings to ensure that the methods used perform reliably in real-world scenarios.
\end{itemize}
For a fair comparison, all DRL models were trained on a Linux server with 92 CPU cores (Intel\textregistered{} Xeon\textregistered{} Platinum 8468V), 922\, GB of RAM, and a single NVIDIA GeForce RTX 4090D GPU. All methods were subsequently tested on a Linux server with 32 CPU cores and 120\, GB of RAM.

\subsection{Deployment System}
\label{sec: deployment_system}
\begin{figure}[htbp]
  \centering
  \includegraphics[width=0.8\linewidth]{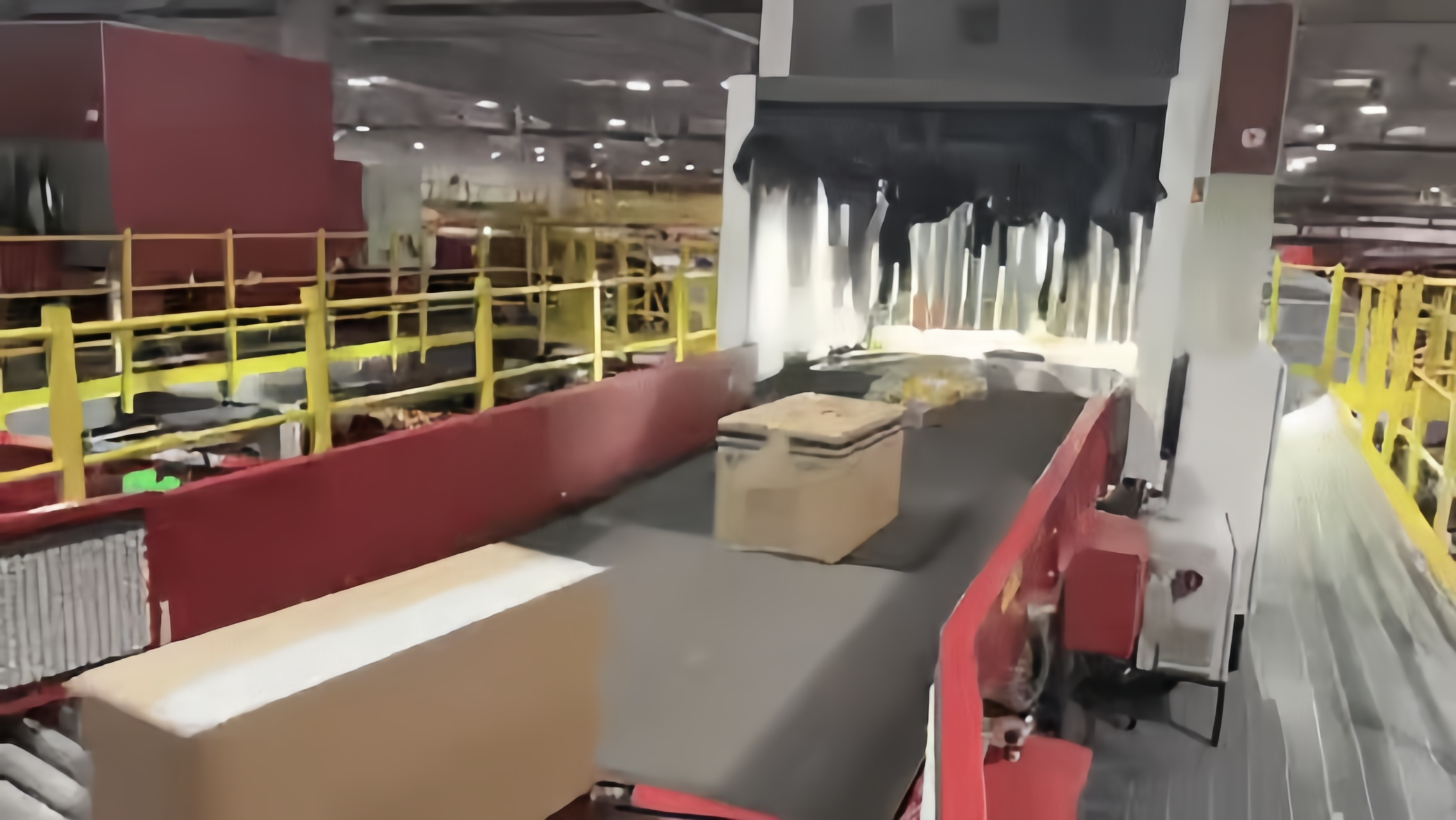}
  \caption{The scanner to upload parcel information onto the Cyber Physical System.}
  \label{fig: deployment_system_scanner}
\end{figure}
\begin{figure}[htbp]
  \centering
  \includegraphics[width=0.8\linewidth]{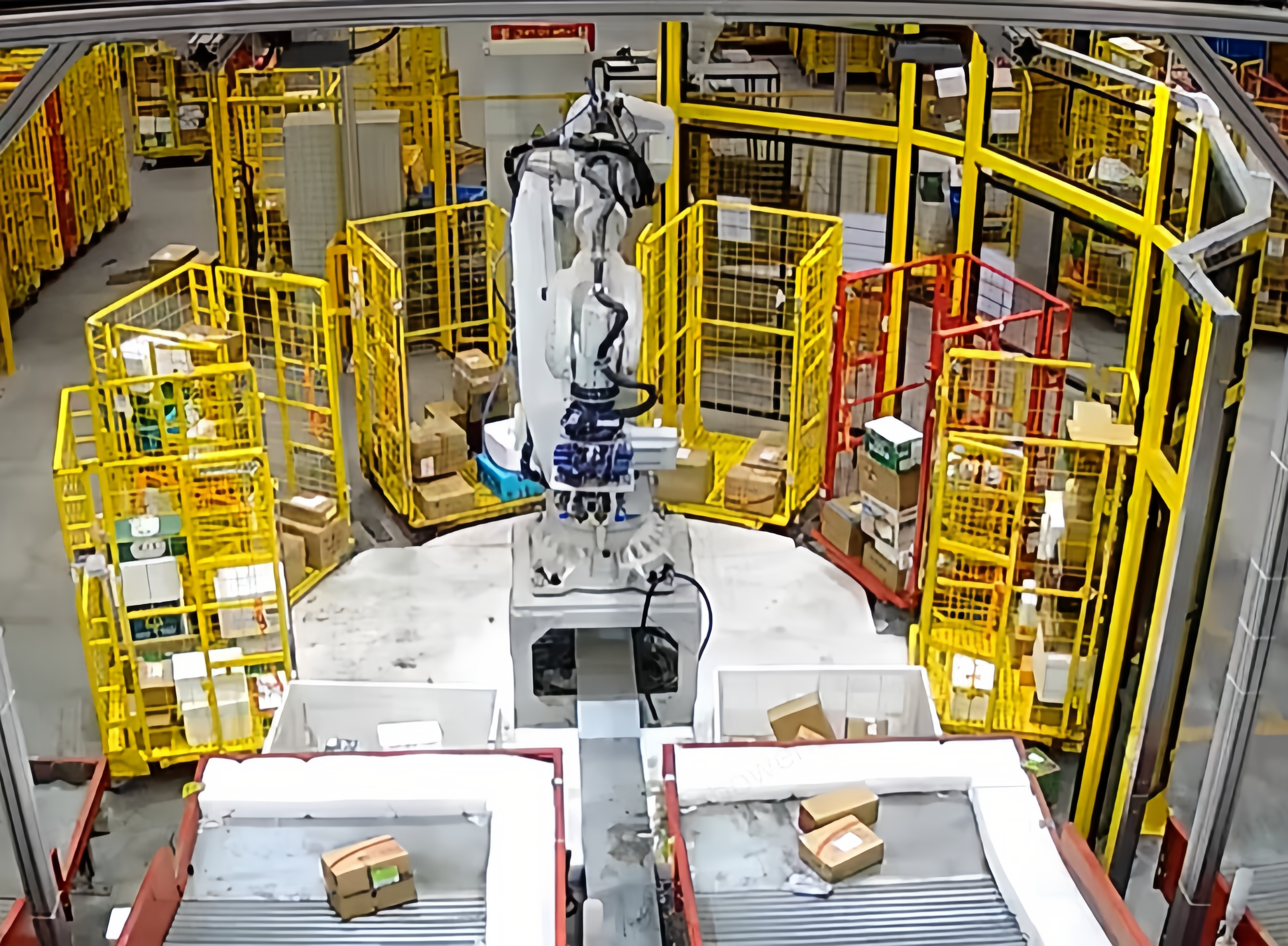}
  \caption{The overview of online 3D-BP with robotic arms.}
  \label{fig: deployment_system_robotic_arm}
\end{figure}
The intelligent online 3D-BP system at JD Logistics can process at least 1500 parcels per robotic arm daily, reducing labor costs by the equivalent of approximately one full-time worker per robotic arm. The system consists of a scanner to obtain parcel information and a conveyor system for routing. As shown in Figure~\ref{fig: deployment_system_scanner}, the scanner reads a parcel's barcode from any face as it passes through. This information is then uploaded to the Cyber-Physical System, which determines the parcel's destination route. When the parcel reaches the end of the conveyor belt (Figure~\ref{fig: deployment_system_robotic_arm}), a top-down camera captures an image and transmits it to the system. Our system uses a YOLO-based model~\cite{jiang2022review} to detect the parcel in the image and calculates some suitable grasp points. Additionally, the server determines the optimal placement location by feeding the current bin state to an online DRL-based 3D-BP model. Finally, the central server computes a collision-free trajectory from the initial position of parcels to the target placement using CuRobo~\cite{sundaralingam2023curobo}, avoiding the bin walls and other packed parcels.
\end{document}